\documentclass[10pt,twocolumn,letterpaper]{article}
\pdfoutput=1

\font\elvbf  = ptmb scaled 1100

\def\abstract
  {%
  \centerline{\large\bf Abstract}%
  \it%
  }

\makeatletter
\DeclareRobustCommand\onedot{\futurelet\@let@token\@onedot}
\def\@onedot{\ifx\@let@token.\else.\null\fi\xspace}

\def\eg{\emph{e.g}\onedot} 
\def\ie{\emph{i.e}\onedot}

\makeatother

\makeatletter
\newlength{\@ctmp}
\newlength{\@figindent}
\long\def\@makecaption#1#2{
  \setbox\@tempboxa\hbox{\small \noindent #1.~#2}
  \setlength{\@ctmp}{\hsize}
  \addtolength{\@ctmp}{-\@figindent}\addtolength{\@ctmp}{-\@figindent}
  \ifdim \wd\@tempboxa >\@ctmp
      {\small #1.~#2\par}
  \else
      \hbox to\hsize{\hfil\box\@tempboxa\hfil}
  \fi}

\def\iccvsection{\@startsection {section}{1}{\z@}
  {10pt plus 2pt minus 2pt}{7pt} {\large\bf}}
\def\iccvssect#1{\iccvsection*{#1}}
\def\iccvsect#1{\iccvsection{\hskip -1em.~#1}}
\def\section{\@ifstar\iccvssect\iccvsect}

\def\iccvsubsection{\@startsection {subsection}{2}{\z@}
  {8pt plus 2pt minus 2pt}{6pt} {\elvbf}}
\def\iccvssubsect#1{\iccvsubsection*{#1}}
\def\iccvsubsect#1{\iccvsubsection{\hskip -1em.~#1}}
\def\subsection{\@ifstar\iccvssubsect\iccvsubsect}
\makeatother

\usepackage{times}
\usepackage{epsfig}
\usepackage{graphicx}
\usepackage{amsmath}
\usepackage{amssymb}
\usepackage{microtype}
\usepackage{booktabs} 
\usepackage{url}            
\usepackage{nicefrac}       

\usepackage[british]{babel}

\usepackage[numbers]{natbib}
\graphicspath{{images/}}

\usepackage{pgfplots}
\pgfplotsset{compat=1.17}
\usepackage{tikzscale}

\usepackage[ruled,vlined]{algorithm2e}

\usepackage{tabularx}
\usepackage{tabulary}
\usepackage{multirow}
\newcolumntype{Y}{>{\centering\arraybackslash}X}

\usepackage{caption}
\usepackage{subcaption}

\usepackage{amsmath,amsfonts,bm}

\DeclareMathOperator*{\softmax}{softmax}
\DeclareMathOperator*{\smoothmax}{smoothmax}

\DeclareMathOperator*{\mindisttolineseg}{d^2_{seg}}
\DeclareMathOperator*{\mindisttocurve}{d^2_{cur}}
\DeclareMathOperator{\softor}{c_{softor}}

\DeclareMathOperator*{\erf}{erf}

\newcommand{\p}{p}
\newcommand{\q}{q}
\newcommand{\probsym}{\q}

\newcommand{\true}[1]{
\global\let\probsym\p
#1
\global\let\probsym\q
}

\usepackage{amssymb}
\usepackage{mathtools}

\usepackage{bm}
\usepackage{chngcntr}
\usepackage{float}
\usepackage{stfloats} 
\usepackage{xspace}

\SetKwProg{Fn}{Function}{}{end}
\SetKwFunction{FCurveDistRecurs}{MinDistanceToCurveBruteForce}%
\SetKwFunction{FCurveDistPolyline}{MinDistanceToCurvePolyline}%

\usepackage{adjustbox}
\usepackage{diagbox}
\usepackage[inline]{enumitem}

\usepackage{xr}
\usepackage{xr-hyper}

\usepackage[pagebackref=true,breaklinks=true,colorlinks,bookmarks=false]{hyperref}

\usepackage{cleveref}

\makeatletter
\renewcommand{\paragraph}{%
  \@startsection{paragraph}{4}%
  {\z@}{1\baselineskip \@plus 1ex \@minus .2ex}{-1em}%
  {\normalfont\normalsize\bfseries}%
}
\makeatother

\usepackage{flushend}

\def\ie{\emph{i.e}\onedot} 
\usepackage{titling}

\date{}

\begin{document}

\title{Differentiable Drawing and Sketching}

\author{Daniela Mihai\footnotemark[1] \\
The University of Southampton \\
Southampton, UK\\
{\tt\small adm1g15@ecs.soton.ac.uk}
\and
Jonathon Hare\footnotemark[1] \\
The University of Southampton \\
Southampton, UK\\
{\tt\small jsh2@ecs.soton.ac.uk}
}

\twocolumn[{%
\renewcommand\twocolumn[1][]{#1}%
\maketitle%
\vspace{-3em}%
\begin{center}
    \centering
    \captionsetup{type=figure}
    \begin{subfigure}[b]{0.31\textwidth}
        \centering
        \resizebox{0.9\textwidth}{!}{
\tikzset {_f6i176qmf/.code = {\pgfsetadditionalshadetransform{ \pgftransformshift{\pgfpoint{-12.5 bp } { 12.5 bp }  }  \pgftransformscale{1 }  }}}%
\pgfdeclareradialshading{_2qpmttutb}{\pgfpoint{0bp}{0bp}}{rgb(0bp)=(1,1,1);%
rgb(0bp)=(1,1,1);%
rgb(5.625bp)=(0,0,0);%
rgb(400bp)=(0,0,0)}%
\tikzset {_esra692vu/.code = {\pgfsetadditionalshadetransform{ \pgftransformshift{\pgfpoint{-12.5 bp } { 12.5 bp }  }  \pgftransformscale{1 }  }}}%
\pgfdeclareradialshading{_mspbzepy0}{\pgfpoint{0bp}{0bp}}{rgb(0bp)=(1,1,1);%
rgb(0bp)=(1,1,1);%
rgb(5.625bp)=(0,0,0);%
rgb(400bp)=(0,0,0)}%
\tikzset {_wwsn9pt49/.code = {\pgfsetadditionalshadetransform{ \pgftransformshift{\pgfpoint{-12.5 bp } { 12.5 bp }  }  \pgftransformscale{1 }  }}}%
\pgfdeclareradialshading{_ck6d22s2t}{\pgfpoint{0bp}{0bp}}{rgb(0bp)=(1,1,1);%
rgb(0bp)=(1,1,1);%
rgb(5.625bp)=(0,0,0);%
rgb(400bp)=(0,0,0)}%
\tikzset {_a0mgku6zf/.code = {\pgfsetadditionalshadetransform{ \pgftransformshift{\pgfpoint{-12.5 bp } { 12.5 bp }  }  \pgftransformscale{1 }  }}}%
\pgfdeclareradialshading{_72tmkbv2s}{\pgfpoint{0bp}{0bp}}{rgb(0bp)=(1,1,1);%
rgb(0bp)=(1,1,1);%
rgb(5.625bp)=(0,0,0);%
rgb(400bp)=(0,0,0)}%
\tikzset{every picture/.style={line width=0.75pt}} 
\begin{tikzpicture}[x=0.75pt,y=0.75pt,yscale=-1,xscale=1]%
\draw  [color={rgb, 255:red, 155; green, 155; blue, 155 }  ,draw opacity=1 ][fill={rgb, 255:red, 255; green, 255; blue, 255 }  ,fill opacity=1 ] (162.11,67.11) -- (200,67.11) -- (200,105) -- (162.11,105) -- cycle ;
\draw [fill={rgb, 255:red, 255; green, 255; blue, 255 }  ,fill opacity=1 ]   (171.58,90.79) .. controls (171.82,100.74) and (190.53,100.74) .. (190.53,90.79) ;
\draw  [color={rgb, 255:red, 155; green, 155; blue, 155 }  ,draw opacity=1 ] (159.74,64.74) -- (197.63,64.74) -- (197.63,102.63) -- (159.74,102.63) -- cycle ;
\draw    (178.68,88.42) -- (178.68,78.95) ;
\draw  [color={rgb, 255:red, 155; green, 155; blue, 155 }  ,draw opacity=1 ] (157.37,62.37) -- (195.26,62.37) -- (195.26,100.26) -- (157.37,100.26) -- cycle ;
\draw   (185.32,72.08) .. controls (185.32,71.95) and (185.42,71.84) .. (185.55,71.84) .. controls (185.68,71.84) and (185.79,71.95) .. (185.79,72.08) .. controls (185.79,72.21) and (185.68,72.32) .. (185.55,72.32) .. controls (185.42,72.32) and (185.32,72.21) .. (185.32,72.08) -- cycle ;
\draw  [color={rgb, 255:red, 155; green, 155; blue, 155 }  ,draw opacity=1 ] (155,60) -- (192.89,60) -- (192.89,97.89) -- (155,97.89) -- cycle ;
\draw   (164.47,69.71) .. controls (164.47,69.58) and (164.58,69.47) .. (164.71,69.47) .. controls (164.84,69.47) and (164.95,69.58) .. (164.95,69.71) .. controls (164.95,69.84) and (164.84,69.95) .. (164.71,69.95) .. controls (164.58,69.95) and (164.47,69.84) .. (164.47,69.71) -- cycle ;
\draw  [color={rgb, 255:red, 155; green, 155; blue, 155 }  ,draw opacity=1 ][fill={rgb, 255:red, 255; green, 255; blue, 255 }  ,fill opacity=1 ] (82.11,97.11) -- (120,97.11) -- (120,135) -- (82.11,135) -- cycle ;
\draw [color={rgb, 255:red, 155; green, 155; blue, 155 }  ,draw opacity=1 ][fill={rgb, 255:red, 255; green, 255; blue, 255 }  ,fill opacity=1 ]   (91.58,120.79) .. controls (91.82,130.74) and (110.53,130.74) .. (110.53,120.79) ;
\draw  [color={rgb, 255:red, 155; green, 155; blue, 155 }  ,draw opacity=1 ][fill={rgb, 255:red, 255; green, 255; blue, 255 }  ,fill opacity=1 ] (79.74,94.74) -- (117.63,94.74) -- (117.63,132.63) -- (79.74,132.63) -- cycle ;
\draw [color={rgb, 255:red, 155; green, 155; blue, 155 }  ,draw opacity=1 ][fill={rgb, 255:red, 255; green, 255; blue, 255 }  ,fill opacity=1 ]   (98.68,118.42) -- (98.68,108.95) ;
\draw  [color={rgb, 255:red, 155; green, 155; blue, 155 }  ,draw opacity=1 ][fill={rgb, 255:red, 255; green, 255; blue, 255 }  ,fill opacity=1 ] (77.37,92.37) -- (115.26,92.37) -- (115.26,130.26) -- (77.37,130.26) -- cycle ;
\draw  [color={rgb, 255:red, 155; green, 155; blue, 155 }  ,draw opacity=1 ][fill={rgb, 255:red, 255; green, 255; blue, 255 }  ,fill opacity=1 ] (105.32,102.08) .. controls (105.32,101.95) and (105.42,101.84) .. (105.55,101.84) .. controls (105.68,101.84) and (105.79,101.95) .. (105.79,102.08) .. controls (105.79,102.21) and (105.68,102.32) .. (105.55,102.32) .. controls (105.42,102.32) and (105.32,102.21) .. (105.32,102.08) -- cycle ;
\draw  [color={rgb, 255:red, 155; green, 155; blue, 155 }  ,draw opacity=1 ][fill={rgb, 255:red, 255; green, 255; blue, 255 }  ,fill opacity=1 ] (75,90) -- (112.89,90) -- (112.89,127.89) -- (75,127.89) -- cycle ;
\draw  [fill={rgb, 255:red, 255; green, 255; blue, 255 }  ,fill opacity=1 ] (84.47,99.71) .. controls (84.47,99.58) and (84.58,99.47) .. (84.71,99.47) .. controls (84.84,99.47) and (84.95,99.58) .. (84.95,99.71) .. controls (84.95,99.84) and (84.84,99.95) .. (84.71,99.95) .. controls (84.58,99.95) and (84.47,99.84) .. (84.47,99.71) -- cycle ;
\draw  [color={rgb, 255:red, 155; green, 155; blue, 155 }  ,draw opacity=1 ][fill={rgb, 255:red, 255; green, 255; blue, 255 }  ,fill opacity=1 ] (150,0) -- (200,0) -- (200,50) -- (150,50) -- cycle ;
\draw    (162.5,25) .. controls (162.81,38.13) and (187.5,38.13) .. (187.5,25) ;
\draw    (175,25) -- (175,12.5) ;
\draw   (162.5,12.81) .. controls (162.5,12.64) and (162.64,12.5) .. (162.81,12.5) .. controls (162.99,12.5) and (163.13,12.64) .. (163.13,12.81) .. controls (163.13,12.99) and (162.99,13.13) .. (162.81,13.13) .. controls (162.64,13.13) and (162.5,12.99) .. (162.5,12.81) -- cycle ;
\draw   (186.88,12.81) .. controls (186.88,12.64) and (187.01,12.5) .. (187.19,12.5) .. controls (187.36,12.5) and (187.5,12.64) .. (187.5,12.81) .. controls (187.5,12.99) and (187.36,13.13) .. (187.19,13.13) .. controls (187.01,13.13) and (186.88,12.99) .. (186.88,12.81) -- cycle ;
\path  [shading=_2qpmttutb,_f6i176qmf] (7.11,68.11) -- (45,68.11) -- (45,106) -- (7.11,106) -- cycle ; 
 \draw  [color={rgb, 255:red, 155; green, 155; blue, 155 }  ,draw opacity=1 ] (7.11,68.11) -- (45,68.11) -- (45,106) -- (7.11,106) -- cycle ; 
\path  [shading=_mspbzepy0,_esra692vu] (4.74,65.74) -- (42.63,65.74) -- (42.63,103.63) -- (4.74,103.63) -- cycle ; 
 \draw  [color={rgb, 255:red, 155; green, 155; blue, 155 }  ,draw opacity=1 ] (4.74,65.74) -- (42.63,65.74) -- (42.63,103.63) -- (4.74,103.63) -- cycle ; 
\path  [shading=_ck6d22s2t,_wwsn9pt49] (2.37,63.37) -- (40.26,63.37) -- (40.26,101.26) -- (2.37,101.26) -- cycle ; 
 \draw  [color={rgb, 255:red, 155; green, 155; blue, 155 }  ,draw opacity=1 ] (2.37,63.37) -- (40.26,63.37) -- (40.26,101.26) -- (2.37,101.26) -- cycle ; 
\path  [shading=_72tmkbv2s,_a0mgku6zf] (0,61) -- (37.89,61) -- (37.89,98.89) -- (0,98.89) -- cycle ; 
 \draw  [color={rgb, 255:red, 155; green, 155; blue, 155 }  ,draw opacity=1 ] (0,61) -- (37.89,61) -- (37.89,98.89) -- (0,98.89) -- cycle ; 
\draw [color={rgb, 255:red, 208; green, 2; blue, 27 }  ,draw opacity=1 ]   (25,45) .. controls (34.19,62.89) and (74.11,48.93) .. (51.1,83.39) ;
\draw [shift={(50,85)}, rotate = 304.76] [color={rgb, 255:red, 208; green, 2; blue, 27 }  ,draw opacity=1 ][line width=0.75]    (10.93,-3.29) .. controls (6.95,-1.4) and (3.31,-0.3) .. (0,0) .. controls (3.31,0.3) and (6.95,1.4) .. (10.93,3.29)   ;
\draw [color={rgb, 255:red, 208; green, 2; blue, 27 }  ,draw opacity=1 ]   (50,90) .. controls (65.1,102.31) and (36.22,107.35) .. (68.48,119.44) ;
\draw [shift={(70,120)}, rotate = 199.82999999999998] [color={rgb, 255:red, 208; green, 2; blue, 27 }  ,draw opacity=1 ][line width=0.75]    (10.93,-3.29) .. controls (6.95,-1.4) and (3.31,-0.3) .. (0,0) .. controls (3.31,0.3) and (6.95,1.4) .. (10.93,3.29)   ;
\draw [color={rgb, 255:red, 208; green, 2; blue, 27 }  ,draw opacity=1 ]   (125,115) .. controls (139.45,104) and (111.53,100.6) .. (148.28,85.69) ;
\draw [shift={(150,85)}, rotate = 518.49] [color={rgb, 255:red, 208; green, 2; blue, 27 }  ,draw opacity=1 ][line width=0.75]    (10.93,-3.29) .. controls (6.95,-1.4) and (3.31,-0.3) .. (0,0) .. controls (3.31,0.3) and (6.95,1.4) .. (10.93,3.29)   ;
\draw [color={rgb, 255:red, 208; green, 2; blue, 27 }  ,draw opacity=1 ]   (150,75) .. controls (126.2,66.8) and (132.19,62.16) .. (143.72,51.22) ;
\draw [shift={(145,50)}, rotate = 496.19] [color={rgb, 255:red, 208; green, 2; blue, 27 }  ,draw opacity=1 ][line width=0.75]    (10.93,-3.29) .. controls (6.95,-1.4) and (3.31,-0.3) .. (0,0) .. controls (3.31,0.3) and (6.95,1.4) .. (10.93,3.29)   ;
\draw (2,3) node [anchor=north west][inner sep=0.75pt]  [font=\scriptsize,color={rgb, 255:red, 144; green, 19; blue, 254 }  ,opacity=1 ] [align=left] {point(-0.5,-0.5)\\point(0.5,-0.5)\\line(0.5,0,0,0)\\bezier(-0.5,0,-0.5,-0.5,0.5,-0.5,0.5.0)};
\draw (115,2) node [anchor=north east] [inner sep=0.75pt]  [font=\scriptsize] [align=left] {Parametric\\Primitives};
\draw (1,108) node [anchor=north west][inner sep=0.75pt]  [font=\scriptsize] [align=left] {Distance\\Transforms};
\draw (66,63) node [anchor=north west][inner sep=0.75pt]  [font=\scriptsize] [align=left] {Relaxed\\Rasterisation};
\draw (199.63,108) node [anchor=north east] [inner sep=0.75pt]  [font=\scriptsize] [align=left] {Differentiable\\Composition};
\end{tikzpicture}%
}
        \caption{Overview of the proposed approach to drawing. The final image is differentiable with respect to the primitive parameters.}
        \label{fig:rast-overview}
    \end{subfigure}\hspace{1em}%
    \begin{subfigure}[b]{0.31\textwidth}
        \centering
        \includegraphics[width=0.9\textwidth]{images/imageopt/bathers.pdf}
        \caption{Seurat's `Une baignade à Asnières' reduced to straight lines instead of points by gradient descent through the rasteriser.}
        \label{fig:bathers}
    \end{subfigure}
    \hspace{1em}%
    \begin{subfigure}[b]{0.31\textwidth}
        \centering
        \resizebox{\textwidth}{!}{\tikzset{every picture/.style={line width=0.75pt}} 
\begin{tikzpicture}[x=0.75pt,y=0.75pt,yscale=-1,xscale=1]%
\draw (40,40.29) node  {\includegraphics[width=60pt,height=60pt]{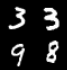}};
\draw (40,130) node  {\includegraphics[width=60pt,height=60pt]{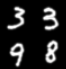}};
\draw  [fill={rgb, 255:red, 245; green, 166; blue, 35 }  ,fill opacity=1 ] (100.41,0.84) -- (159.82,17.78) -- (160.17,61.55) -- (101.04,80.17) -- cycle ;
\draw  [fill={rgb, 255:red, 255; green, 255; blue, 255 }  ,fill opacity=1 ] (210,0.29) -- (210,80.29) -- (180,80.29) -- (180,0.29) -- cycle ;
\draw    (90,40.29) -- (80,40.29) -- (97,40.29) ;
\draw [shift={(100,40.29)}, rotate = 180] [fill={rgb, 255:red, 0; green, 0; blue, 0 }  ][line width=0.08]  [draw opacity=0] (6.25,-3) -- (0,0) -- (6.25,3) -- cycle    ;
\draw    (172.83,40.29) -- (160,40.29) -- (177.43,40.49) ;
\draw [shift={(180.43,40.53)}, rotate = 180.67] [fill={rgb, 255:red, 0; green, 0; blue, 0 }  ][line width=0.08]  [draw opacity=0] (6.25,-3) -- (0,0) -- (6.25,3) -- cycle    ;
\draw    (230.76,40.29) -- (211.52,40.29) -- (227.76,40.29) ;
\draw [shift={(230.76,40.29)}, rotate = 180] [fill={rgb, 255:red, 0; green, 0; blue, 0 }  ][line width=0.08]  [draw opacity=0] (6.25,-3) -- (0,0) -- (6.25,3) -- cycle    ;
\draw    (260,60) -- (260,130) -- (213,130) ;
\draw [shift={(210,130)}, rotate = 360] [fill={rgb, 255:red, 0; green, 0; blue, 0 }  ][line width=0.08]  [draw opacity=0] (6.25,-3) -- (0,0) -- (6.25,3) -- cycle    ;
\draw    (180,130) -- (163,130) ;
\draw [shift={(160,130)}, rotate = 360] [fill={rgb, 255:red, 0; green, 0; blue, 0 }  ][line width=0.08]  [draw opacity=0] (6.25,-3) -- (0,0) -- (6.25,3) -- cycle    ;
\draw  [color={rgb, 255:red, 0; green, 0; blue, 0 }  ,draw opacity=1 ][fill={rgb, 255:red, 226; green, 189; blue, 234 }  ,fill opacity=1 ] (100,90.32) -- (160,107.84) -- (160,152) -- (100,170.48) -- cycle ;
\draw    (100,130) -- (83,130) ;
\draw [shift={(80,130)}, rotate = 360] [fill={rgb, 255:red, 0; green, 0; blue, 0 }  ][line width=0.08]  [draw opacity=0] (6.25,-3) -- (0,0) -- (6.25,3) -- cycle    ;
\draw   (180,90) -- (210,90) -- (210,170) -- (180,170) -- cycle ;
\draw  [fill={rgb, 255:red, 245; green, 166; blue, 35 }  ,fill opacity=1 ] (290,80.11) -- (230,62.47) -- (230,18.47) -- (290,0.11) -- cycle ;
\draw (109,40.29) node [anchor=west] [inner sep=0.75pt]  [font=\small] [align=left] {\begin{minipage}[lt]{30.589375pt}\setlength\topsep{0pt}
\begin{center}
Encoder\\Net
\end{center}
\end{minipage}};
\draw (195.5,63.29) node [anchor=west] [inner sep=0.75pt]  [font=\small,rotate=-270] [align=left] {\begin{minipage}[lt]{33.915pt}\setlength\topsep{0pt}
\begin{center}
Vector\\Encoding
\end{center}
\end{minipage}};
\draw (232.76,40.29) node [anchor=west] [inner sep=0.75pt]  [font=\small] [align=left] {\begin{minipage}[lt]{37.644375000000004pt}\setlength\topsep{0pt}
\begin{center}
Parameter\\Decoder \\Net
\end{center}
\end{minipage}};
\draw (104,130) node [anchor=west] [inner sep=0.75pt]  [font=\small] [align=left] {\begin{minipage}[lt]{36.805pt}\setlength\topsep{0pt}
\begin{center}
Rasteriser
\end{center}
\end{minipage}};
\draw (195.5,159) node [anchor=west] [inner sep=0.75pt]  [font=\small,rotate=-270] [align=left] {\begin{minipage}[lt]{41.384375000000006pt}\setlength\topsep{0pt}
\begin{center}
Primitive\\Parameters
\end{center}
\end{minipage}};
\end{tikzpicture}%
}
        \caption{Encoder-Decoder-Rasteriser model for learning to map images to primitives. Orange blocks have learnable parameters.}
        \label{fig:ae}
    \end{subfigure}
    \caption{With a differentiable rasteriser (a), it is possible to optimise primitives (b), and build end-to-end learnable models (c).}
\end{center}%
}]
{
  \renewcommand{\thefootnote}%
    {\fnsymbol{footnote}}
  \footnotetext[1]{Authors contributed equally.}
}

\begin{abstract}
We present a bottom-up differentiable relaxation of the process of drawing points, lines and curves into a pixel raster. Our approach arises from the observation that rasterising a pixel in an image given parameters of a primitive can be reformulated in terms of the primitive's distance transform, and then relaxed to allow the primitive's parameters to be learned. This relaxation allows end-to-end differentiable programs and deep networks to be learned and optimised, and provides several building blocks that allow control over how a compositional drawing process is modelled. We emphasise the bottom-up nature of our proposed approach, which allows for drawing operations to be composed in ways that can mimic the physical reality of drawing rather than being tied to, for example, approaches in modern computer graphics.
With the proposed approach we demonstrate how sketches can be generated by directly optimising against photographs and how auto-encoders can be built to transform rasterised handwritten digits into vectors without supervision. Extensive experimental results highlight the power of this approach under different modelling assumptions for drawing tasks.
\end{abstract}

\section{Introduction and Motivation}

This paper proposes a differentiable relaxation of the rasterisation process, which we ultimately demonstrate allows us to build end-to-end learnable machines that can perform both image generation and inference tasks. More concretely, we demonstrate that we can build machines that turn digital raster images into parametric representations of continuous paths, and back into rasterised images again. Our approach is not constrained by any particular modelling assumptions about how images should be composed beyond the functions used being differentiable. This allows us to closely model the physical act of drawing with a pen on paper for example. We believe that the proposed approach will ultimately have many applications in future approaches to computer vision tasks related to topics including sketch retrieval, recognition, and generation, as well as topics related to understanding and analysing handwriting, and even to more general topics around understanding visual communication.

When humans use drawing, sketching and writing to communicate they rarely do so by filling in pixels on a grid. Most methods (with some notable exceptions
) of producing physically realised forms of drawing and writing by hand involve manipulating an instrument (a pen, paintbrush, pastel, etc) to mark a surface (paper, for example). In the digital world, this process is often approximated with vector graphics, in which paths are `stroked' and then most often rasterised onto a pixel grid to produce digital images that can be displayed on a monitor or reproduced in hard copy. 

To date, modelling the act of drawing with techniques such as deep neural networks has been relatively limited because the process of rasterisation using traditional approaches is not differentiable. The vast majority of recent work on image generation has operated on the principle of trying to optimise outputs broadly at the pixel level utilising tools such as transpose convolutions which operate on raster representations. There are of course exceptions to this statement, where researchers have attempted to more closely consider the underlying process that humans use to draw and write~\citep[\eg][]{lake2015human}, or to circumvent the non-differentiability of rasterisation \citep[\eg][]{zheng2018strokenet} using learning. These techniques, as well as a contemporaneous approach to relaxing modern vector graphics~\citep{10.1145/3414685.3417871} (taking a complementary, but different approach to ours) which was published during the production of this manuscript, are described and discussed in \cref{sec:related}.

Our contributions are as follows:
\begin{enumerate*}
    \item We present a bottom-up differentiable approach (see \cref{fig:rast-overview}) to generating pixel rasters from parameterised vector primitives by reformulating and relaxing the rasterisation problem. This is coupled with a set of formulations that allow different approaches to composition. A full exposition of our approach is in \cref{sec:relaxedrast}.
    \item We demonstrate that primitives can be optimised by minimising a loss against an existing raster image (cf. \cref{fig:bathers}), and show how different losses inform the result. Details are in \cref{sec:optimisation}.
    \item We create a range of parameterisations of primitives in end-to-end learnable autoencoder architectures (see \cref{fig:ae}) for handwritten characters and objectively compare performance. See \cref{sec:ae} for details.
    We provide a PyTorch implementation of our approach, which allows others to experiment further. Code is available at \url{https://bit.ly/2PHtt5v}.
\end{enumerate*}


\section{Related Work}
\label{sec:related}
Drawing, and in particular sketching, has been a means of conveying concepts, objects and stories since ancient times. There is a long history of sketch research in computer vision and human-computer interaction dating back to the 1960s~\citep{sutherland1964sketchpad}. Sketch applications have become increased in recent years due to the rapid development of deep-learning techniques that can successfully tackle tasks such as sketch recognition~\citep{yu2017sketch}, generation~\citep{zheng2018strokenet, ha2018neural, sangkloy2017scribbler}, sketch-based retrieval~\citep{choi2019sketchhelper, sangkloy2016sketchy, creswell2016adversarial}, semantic segmentation~\citep{Sketchsegnet, yang2020sketchgcn}, grouping\citep{Li_2018_ECCV}, parsing\citep{2017sketchparse} and abstraction\citep{Muhammad_2018_CVPR}. \citet{xu2020deep} offer a recent and detailed survey of free-hand sketch research and applications, focusing on contemporary deep-learning techniques. Our long-term goal for the work presented in this paper is to be able to train models to learn how to produce the parameters of drawing primitives based on visual inputs with only limited supervision. Internally within our models, we want to bridge the gap between input and output rasters, and internal vector representations. 

There is a body of recent literature describing models that operate purely on vector stroke data (that is, the process of actually drawing the vectors into an image is not part of the learning machinery). This includes recurrent generative models for sketch data~\citep[\eg][]{ha2018neural}, generative models utilising GANs for sketch generation in vector format~\citep[\eg][]{balasubramanian2019teaching}, and reinforcement learning~\citep[\eg][]{doodlingRL, xie2013artist}. Another line of work within sketch generation uses Bayesian Program Learning, rather than deep networks, to represent the act of drawing as a probabilistic generative model~\citep{lake2015human}. 

With respect to models that turn raster images into vectors, there is considerable classical literature looking at the problem of `stroke-based rendering' where the objective is to turn raster images into a sequence of strokes~\citep[\eg][]{10.1145/237170.237287,10.1145/280814.280951,10.1145/192161.192184} for artistic or visual communication purposes. A good overview of these can be found in the tutorial by \citet{1210867}, which breaks these approaches into Voronoi (broadly based on Lloyd's algorithm), or `trial and error' approaches which try to minimise a loss based on heuristic tests. The approach presented in \cref{sec:optimisation} is clearly of relevance to this field of research, but a differentiable rasteriser allows for a potentially more principled or flexible approach to modelling the drawing process or the loss that is optimised. A number of models have been proposed that incorporate drawing into learning machinery. Until very recently, the process of rasterisation and rendering was thought to be non-differentiable, so two approaches were used to circumvent this problem: firstly there were models that use reinforcement learning to learn drawing actions through a traditional (non-differentiable) renderer~\citep[\eg][]{DBLP:conf/icml/GaninKBEV18,DBLP:journals/corr/abs-1910-01007}, and, secondly there were approaches that `learn' renderers (typically formulated as networks of transposed convolutions, or convolutions and upsampling operations) that can take vector inputs and produce raster outputs~\citep{zheng2018strokenet,zou2020stylized,DBLP:journals/corr/abs-1904-08410}. Of the latter, the work by \citet{zheng2018strokenet} is most similar to ours in its intent to work with sketches, and to utilise encoder models to produce accurate stroke parameters from raster images; our models in \cref{sec:ae} are however fully end-to-end learnable, unlike \citeauthor{zheng2018strokenet}'s model in which the renderer network is trained separately. Models with learned rasterisers are also inherently inflexible in the sense that they have to be trained for every type of stroke parameterisation they can work with.

Recent approaches to differentiable 3D rendering have garnered attention in the computer vision community~\citep[\eg][]{9008817,kato2018renderer}, and indeed it is the work of \citet{9008817} that originally helped inform the approach we detail in \cref{sec:relaxedrast}. Most recently, during the development of our approach,  \citet{10.1145/3414685.3417871} presented a differentiable relaxation that takes advantage of how anti-aliasing is performed in modern computer graphics systems using multi-sampling, by providing differentiable relaxations. We consider this to be a top-down approach to the problem because it does not change the underlying rendering model. Conversely, we consider our approach to be bottom-up because we explicitly allow the rendering model to be flexibly defined in a way that is appropriate to the task.

\section{Differentiable relaxations of rasterisation}
\label{sec:relaxedrast}

\begin{figure*}[t]
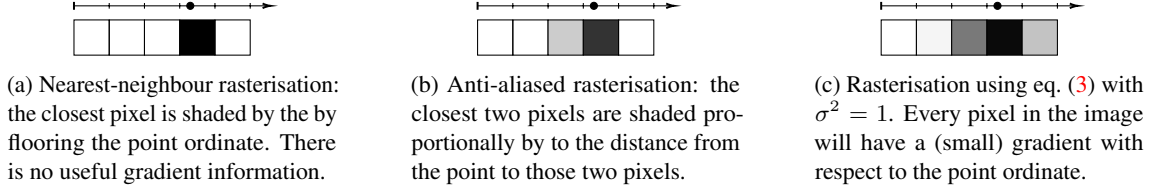

    \centering
    \begin{subfigure}[t]{.25\textwidth}
        \centering
        \includegraphics[width=0.6\textwidth]{nn1d}
        \caption{Nearest-neighbour rasterisation: the closest pixel is shaded by the by flooring the point ordinate.  There is no useful gradient information.}
        \label{fig:nnrast}
    \end{subfigure}\hspace{1cm}%
    \begin{subfigure}[t]{.25\textwidth}
        \centering
        \includegraphics[width=0.6\textwidth]{aa1d}
        \caption{Anti-aliased rasterisation: the closest two pixels are shaded proportionally by to the distance from the point to those two pixels.}
        \label{fig:aarast}
  \end{subfigure}\hspace{1cm}%
  \begin{subfigure}[t]{.25\textwidth}
    \centering
    \includegraphics[width=0.6\textwidth]{exp1d}
    \caption{Rasterisation using \cref{eqn:exp1d} with $\sigma^2 = 1$. Every pixel in the image will have a (small) gradient with respect to the point ordinate.}
    \label{fig:exp1d}
    \end{subfigure}
  \caption{Different point rasterisation functions illustrated in one-dimension.}\label{fig:rasfcns}
\end{figure*}

In this section we discuss the problem of drawing, or rasterising points, lines and curves defined in a continuous world space $\mathcal{W}$ into an image space $\mathcal{I}$. Our objective is to present a formalisation that allows us to ultimately define rasterisation functions that are differentiable with respect to their world space parameters (\eg the (co)ordinate of a point, or (co)ordinates of the beginning and end of a line segment).

\subsection{1D Rasterisation}
We first consider the problem of rasterising a one-dimensional point $p \in \mathcal{W}$ where $\mathcal{W} = \mathbb{R}$. Concretely, the process of rasterisation of the point $p$ can be defined by a function, $f(n; p)$, that computes a value (typically $[0, 1]$) for every pixel in the image space $\mathcal{I}$, whose position is given by $n \in \mathcal{I}$.  Such a function represents a scalar field over the space of possible values of $n$. Commonly we consider values of $n$ to be non-negative integers from the lattice or grid, $\mathbb{Z}^1_{0+}$, defining a pixel in the image. 

\paragraph{Simple closest-pixel rasterisation functions.}
If we assume that the 0th pixel covers the domain $[0,1)$ in the world space of a point $p$, and that the 1st pixel covers $[1,2)$, etc. Nearest-neighbour rasterisation then maps the real-valued point, $p$, to an image by rounding down:
\begin{equation}
    f(n; p) = \begin{cases} 
      1 & \mbox{if } \lfloor p \rfloor = n \\
      0 & \mbox{otherwise} \; .
   \end{cases}
   \label{eqn:nn1d}
\end{equation}
This process is illustrated in \cref{fig:nnrast}. An alternative rasterisation scheme, illustrated in \cref{fig:aarast} is to interpolate over the two closest pixels. Assuming that a pixel has maximal value when the point being rasterised lies at its midpoint, then:
\begin{equation}
    f(n; p) = \begin{cases} 
      1.5 - p + \lfloor p - 0.5 \rfloor & \mbox{if } \lfloor p - 0.5 \rfloor = n \\
      0.5 + p - \lceil p - 0.5 \rceil & \mbox{if } \lceil p - 0.5 \rceil = n \\
      0 & \mbox{otherwise} \; .
   \end{cases}
   \label{eqn:aa1d}
\end{equation}
These functions (extended to 2D) are actually implicitly used in many computer graphics systems, but rarely in the form we have written them. Most graphics subroutines approach the rasterisation problem from the perspective of directly determining which pixels in $n$ should have a colour associated with them given $p$ as this is more efficient if the objective is just to draw the primitive $p$.

\paragraph{Differentiable Relaxations.}
Ideally, we would like to be able to define a rasterisation function that is differentiable with respect to $p$. This would allow $p$ to be optimised with respect to some objective. The rasterisation function given by \cref{eqn:nn1d} is piecewise differentiable with respect to $p$, but the gradient is zero almost everywhere which is not useful. Although \cref{eqn:aa1d} has some gradient in the two pixels nearest to $p$, overall it has the same key problem: the gradient is zero almost everywhere. 

We would like to define a rasterisation function that has gradient for all (or at least a large proportion of) possible values of $n$. This function should be continuous and differentiable almost everywhere. The anti-aliased rasterisation approach gives some hint as to how this could be achieved: the function could compute a value for every $n$ based on the distance between $n$ and $p$. Distance metrics have an infinite upper bound, whereas we want our pixel values to be finitely bounded in $[0,1]$, so inversion and application of a non-linearity are necessary. The properties of the chosen function should give values close to 1 when $n$ and $p$ are \textit{close}, and values near 0 when they are \textit{far apart}.

An obvious choice of non-linearity would be to exponentiate the negative squared distances, and use a scaling factor $\sigma^2$ to control the fuzziness of the rasterisation and the size of the point or width of the line stroke (see \cref{fig:exp1d}):
\begin{equation}
    f(n; p) = \exp(-d^2(n, p - 0.5) / \sigma^2) \; .
    \label{eqn:exp1d}
\end{equation}
It can be shown that there is a direct linear relationship (see proof in \cref{sup:thickness}) between the size of a point or thickness of a line, $t$, and the value of $\sigma$: $\sigma \approx 0.54925 t\,\forall\, t>0$.

\subsection{Relaxed Rasterisation in N-dimensions}
All of the 1D rasterisation functions previously defined can be trivially extended to rasterise a \textit{point} in two or more dimensions. For example, if the point $\bm{p}$ was considered to be a vector in the world space $\mathcal{W} = \mathbb{R}^2$ and correspondingly $\bm{n}$ was a vector in the image space\footnote{Note that it is most common to use positive integers to index pixels in the image space, but this isn't a requirement; the image space could be unbounded or real for example.}, $\mathcal{I} = \mathbb{Z}_{0+}^2$, and the floor and ceiling operators are applied element-wise then all three 1D rasterisation functions hold in two (or more) dimensions.

\paragraph{Line Segments.} A \textit{line segment} can be defined by its start coordinate $\bm{s}=[s_x,s_y]$ and end coordinate $\bm{e}=[e_x,e_y]$. The normal approaches to rasterising lines in computer graphics~\citep[\eg][]{Bresenham1965,XiaolinWu1991} are highly optimised and work by considering just the pixels that intersect the line or are within a few pixels of it. These algorithms typically iterate over the line, setting the underlying pixels values accordingly. To develop a general set of (potentially differentiable) rasterisation functions we need to consider a formalisation of rasterisation as we did in the 1D case where we consider a function that defines a scalar field over the set of all pixel positions, $n$, in the image given a particular line segment:~$f(\bm{n}; \bm{s}, \bm{e})$.

To rasterise a line segment one needs to consider how close a pixel is to the segment. We can efficiently compute the squared Euclidean distance of an arbitrary pixel $\bm{n}$ to the closest point on the line segment as follows:
\begin{flalign} 
\begin{tikzpicture}[overlay]
\node[xshift=5.9cm,yshift=-0.15cm] at (0,0){%
    \includegraphics[width=2cm]{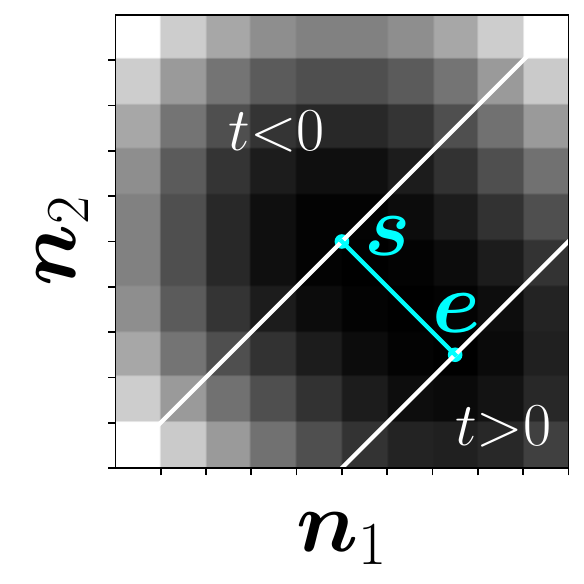}};
\end{tikzpicture}
    \bm{m} & = \bm{e} - \bm{s} \;, \nonumber \\
    t & = ((\bm{n} - \bm{s}) \cdot \bm{m}) / (\bm{m} \cdot \bm{m}) \;, \nonumber \\
    \mindisttolineseg(\bm{n}, \bm{s}, \bm{e}) & = \begin{cases} 
      || \bm{n} - \bm{s} ||_2^2 & \mbox{if } t \leq 0\\
      || \bm{n} - (\bm{s} + t \bm{m}) ||_2^2 & \mbox{if } 0 < t < 1\\
      || \bm{n} - \bm{e} ||_2^2 & \mbox{if } t \geq 1 \; .
   \end{cases}&&
   \label{eqn:closestPointLine}
\end{flalign}
Concretely, $\mindisttolineseg(\bm{n}, \bm{s}, \bm{e})$ is the squared Euclidean Distance Transform of the line segment. It defines a scalar field in which the value is equal to the squared distance to the closest point on the line segment. This function is piecewise smooth and differentiable with respect to the line segment parameters everywhere for a given $\bm{n}$.

In the case of nearest-neighbour rasterisation one would ask if the line passes through the pixel in question and only fill it if that were the case:
\begin{flalign}
\begin{tikzpicture}[overlay]
\begin{scope}[xshift=6.3cm,yshift=0.75cm]
\node at (0,0){%
    \includegraphics[width=0.9cm]{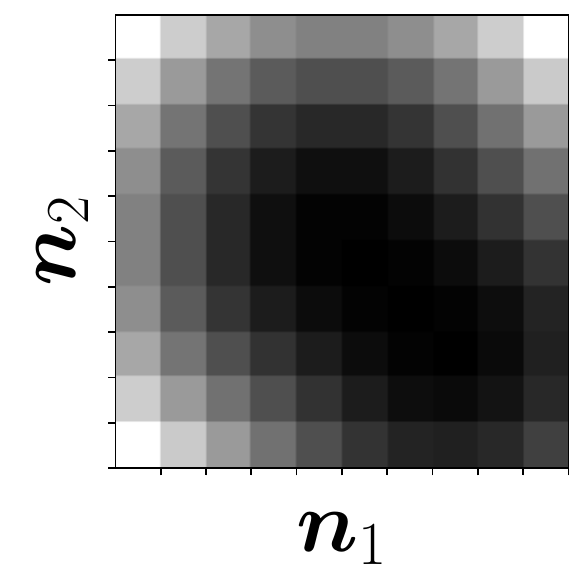}};
\node at (1.2,0){%
    \includegraphics[width=0.9cm]{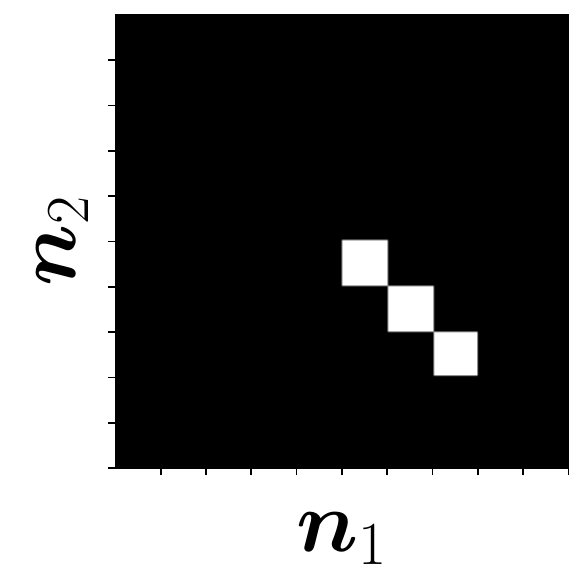}};
    \draw [-stealth](0.5,0.07) -- (0.75,0.07);
\end{scope}
\end{tikzpicture}
    f(\bm{n}; \bm{s}, \bm{e}) &= \begin{cases} 
      1 & \mbox{if } \mindisttolineseg(\bm{n}, \bm{s}, \bm{e}) \leq \delta^2\\
      0 & \mbox{otherwise} \; .
   \end{cases}&&
   \label{eqn:nnLineRast}
\end{flalign}
Assuming a 1-1 mapping between the domains of the coordinate system of the image space and world space, then $\delta^2=0.5$ would give a rasterisation that mimics the 1-pixel wide line that would be drawn by Bresenham's algorithm~\citep{Bresenham1965}. 
If we replace the calculation of distance to a point in \cref{eqn:exp1d} with the minimum distance to the line segment we get a line segment rasteriser that is differentiable with respect to the parameters of the line segment $\bm{s}$ and $\bm{e}$:
\begin{flalign}	
\begin{tikzpicture}[overlay]	
\begin{scope}[xshift=6.3cm,yshift=0.55cm]	
\node at (0,0){%
    \includegraphics[width=0.9cm]{images/equations/eqn4b.pdf}};	
\node at (1.2,0){%
    \includegraphics[width=0.9cm]{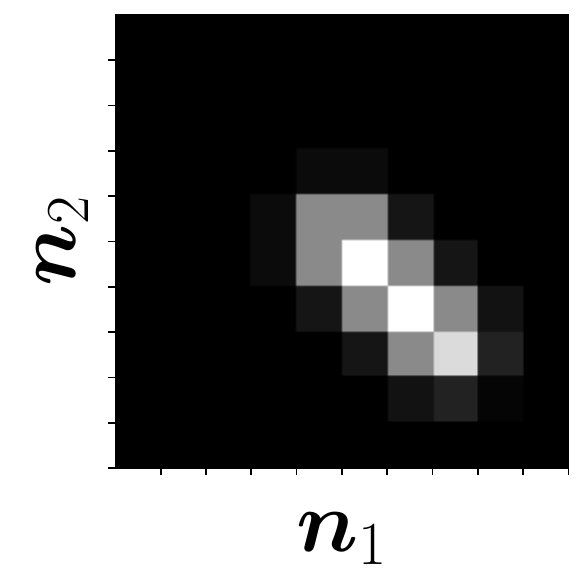}};	
    \draw [-stealth](0.5,0.07) -- (0.75,0.07);	
\end{scope}	
\end{tikzpicture}	
    f(\bm{n}; \bm{s}, \bm{e}) = \exp(-\mindisttolineseg(\bm{n}, \bm{s}, \bm{e}) / \sigma^2) \;. &&	
    \label{eqn:expLine2d}	
\end{flalign}

\paragraph{Curves.}
\label{sec:diffcurverender}
It is common in computer graphics to utilise parametric curves $C(t, \bm\theta)$ where $\bm \theta$ defines the parameters and $0 \leq t \leq 1$. Typically $C(t, \bm \theta)$ is polynomial (usually quadratic or cubic in $t$). The parameters $\bm \theta$ are commonly specified in B\'ezier (\eg B\'ezier Curves) or Hermite form (\eg Catmull-Rom splines) as described in \cref{sup:curveparam}.  To rasterise a curve (irrespective of the parameterisation) in a way that is differentiable with respect to the parameters we can follow the same general approach that was taken for line segments: \textit{compute the minimum Squared Euclidean distance between each coordinate $\bm{n} \in \mathcal{I}$ and the curve}:
\begin{flalign}	
    \begin{tikzpicture}[overlay]	
\node[xshift=7.55cm,yshift=0.55cm] at (0,0){%
    \includegraphics[width=1.5cm]{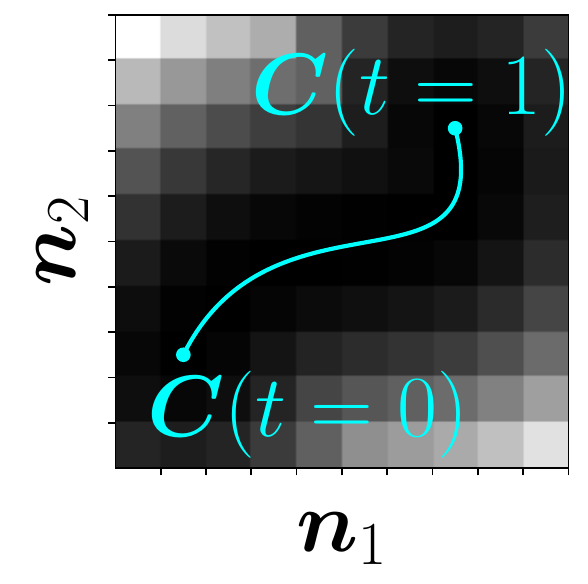}};	
\end{tikzpicture}	
    \rule{1.5cm}{0pt}	
    \begin{aligned}	
        \mindisttocurve(\bm{n}, \bm{\theta}) = \min_{t} \quad & ||C(t, \bm{\theta}) - \bm{n}||^2_2 \\	
        \textrm{s.t.} \quad & 0 \leq t \leq 1 \; .\\	
    \end{aligned}\rule{1.5cm}{0pt}\raisetag{1.3em}	
\end{flalign}
As in the case of line segments, this distance transform can then be combined with a rasterisation function that works in terms of a distance:
\begin{flalign}	
\begin{tikzpicture}[overlay]	
\begin{scope}[xshift=6.1cm,yshift=0.55cm]	
\node at (0,0){%
    \includegraphics[width=0.9cm]{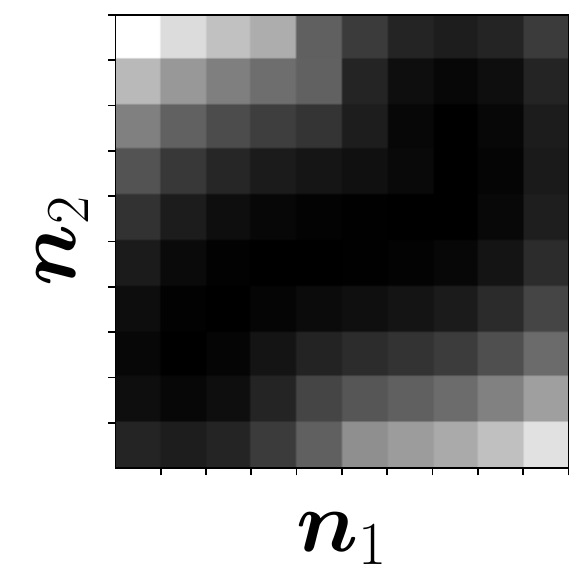}};	
\node at (1.2,0){%
    \includegraphics[width=0.9cm]{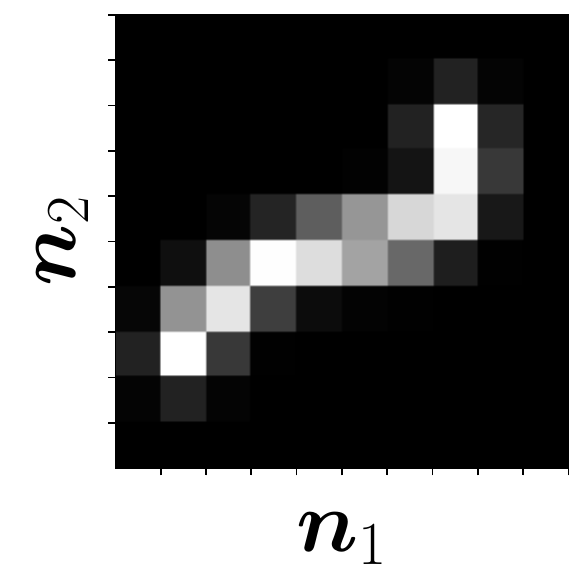}};	
    \draw [-stealth](0.5,0.07) -- (0.75,0.07);	
\end{scope}	
\end{tikzpicture}\;\;\;\;\;\;\;	
    f(\bm{n}; \bm{\theta}) = \exp(-\mindisttocurve(\bm{n}, \bm{\theta}) / \sigma^2) \;.&&	
    \label{eqn:expCurve2d}	
\end{flalign}
The only additional challenge over the rasterisation of line segments is that the computation of the distance map requires solving a constrained minimisation problem and doesn't have a closed form solution. A number of different approaches are possible (see \cref{sup:curveedt}), however, in practice, we have had success with both a fast polyline approximation\footnote{Note that there is a potential for a small change in curve's parameters to cause a large difference in the polyline approximation, although we have not seen this become an issue in practice.} and a recursive approach which are both easily vectorised as tensor operations that can be performed efficiently on a GPU.

\subsection{Composing Multiple Primitives}
\label{ssec:compositions}
To rasterise multiple lines\footnote{We're considering composing multiple line segments, but everything here also applies to multiple points and curves, as well as combinations of line segments, points and curves, or indeed any other raster.} we can consider combining the rasterisations of different line segments into a single image. We denote images produced by rasterising different line segments $\{\bm{s_1},\bm{e_1}\}, \{\bm{s_2},\bm{e_2}\}, \dots, \{\bm{s_i},\bm{e_i}\}$ into matrices $\bm{I}^{(1)}, \bm{I}^{(2)}, \dots, \bm{I}^{(n)}$ defined over the same image space $\mathcal{I}$. In the simplest case, where we have binary rasterisations, we might consider that the logical-or of corresponding pixels would produce the desired effect of selecting any pixels that were shaded in the individual rasterisations as being shaded in the final output:
\begin{equation}
    c(\bm{I}^{(1)}, \bm{I}^{(2)}, \dots, \bm{I}^{(n)}) = \bm{I}^{(1)} \lor \bm{I}^{(2)} \lor \dots \lor \bm{I}^{(n)} \;.
\end{equation}
We can relax this composition to be differentiable and also allow the pixel values to be non binary (but restricted to $[0,1]$) as follows:
\begin{equation}
    \softor(\bm{I}^{(1)}, \bm{I}^{(2)}, \dots, \bm{I}^{(n)}) = \bm{1} - \prod_{i=1}^n (\bm{1} - \bm{I}^{(i)}) \;.
    \label{eqn:softor}
\end{equation}
Effectively if a pixel is `on' in any of the individual images then this will select it as being `on' in the output. This approach treats all the input images as a set; the output will be the same irrespective of the order they appear in. Undoubtedly many other possible differentiable composition functions exist with alternative properties; we propose and discuss a number of these alternatives in \cref{app:compfcns}. The experiments that follow use the above \emph{soft-or} function, with the notable exception of the image in \cref{fig:bathers} which was generated using the \textit{over} composition (see \cref{app:compfcns}) as this is more appropriate for colour images.

\subsection{Extended drawing}
Clearly, at this point, we now have all the components required to construct a basic drawing system. There are however a number of aspects that haven't been considered, including, for example how to draw in colour. As we focus the remainder of the paper on utilising the approach we have already described, discussion of additional extensions related to drawing and rasterisation can be found in \cref{sup:extendrelax}.

\subsection{Advantages and Limitations}
The rasterisation process described in this section in principle allows gradients to flow from every pixel in the image to the parameters of a rendered primitive (note however that in practice this is not the case because of finite numeric precision). This is in contrast to the work of \citet{10.1145/3414685.3417871} where the gradients are limited by the size of the filter. The advantage of our method is that optimisation should be easier with more gradient, however of course this does itself also have disadvantages. Computationally, our approach can be entirely implemented as batched tensor operations (this includes computation of distance transforms for all primitives), so all computation can be performed on the GPU making use of all available processing resources, and unlike \citet{10.1145/3414685.3417871}'s approach, does not involve the CPU for rendering. The disadvantage is that memory usage could be very high, particularly for batches of large images with many primitives (in our original envisaged use case of exploring simple sketching and writing this is not a problem however). One interesting idea to explore in the future would be to utilise sparse tensors to reduce storage requirements by not storing pixels contributing to no value or gradient. Another potential criticism of our approach is that the generated images will be very slightly blurry as a result of the relaxation; again, for our envisaged use case this is not a problem, and it is always possible to use the relaxation for learning/optimisation, and then switch to a regular render for generation at inference time. Finally, we draw attention to the fact that our approach is not restricted to 2D, and can be \eg directly applied to 3D data for voxel rasterisation.

\section{Direct Optimisation of Primitive Parameters}
\label{sec:optimisation}
With the machinery defined in \cref{sec:relaxedrast} it is now possible to define a complete system that takes the parameters describing primitives and rasterises those primitives into an image. If a loss function is introduced in the image space, between the complete rasterised image and a fixed target image, it becomes possible to  compute gradients with respect to the parameters of the primitives that created the rasterised image. Minimising this loss will adapt the underlying primitives to ``shapes'' that best fit the target image. 

A commonly used `reconstruction' loss function for images is the mean squared error between the target and the generated image. We can thus formalise the optimisation problem as,
\begin{equation}
    \min_{\bm{\theta}} \| R(\bm{\theta})  - \bm{T}\|_2^2 \;,
    \label{eqn:mse}
\end{equation}
for a target image, $\bm T$ and rasterisation function $R$ defined over the same image space $\mathcal{I}$. The rasterisation function itself is defined as a composition $c(\dots)$ (see \cref{ssec:compositions}) over $k$ primitives, themselves rasterised by primitive rasterisation functions, $f^{(i)}$ (\eg \cref{eqn:exp1d,eqn:expLine2d,eqn:expCurve2d}, etc.):
\begin{align}
    R(\bm{\theta}) = c(f^{(1)}(\bm{\theta}^{(1)}), f^{(2)}(\bm{\theta}^{(2)}), \dots, f^{(k)}(\bm{\theta}^{(k)})) \nonumber \\
    \mbox{ where } \bm{\theta} = [ \bm{\theta}^{(1)} | \bm{\theta}^{(2)}| \dots| \bm{\theta}^{(k)} ] \;.
\end{align}
If the rasterisation function $R$ is differentiable with respect to $\bm{\theta}$, then the minimisation problem in \cref{eqn:mse} can be solved using gradient descent. Note that the problem is in general non-convex, with potentially many local optima\footnote{The optimisation landscape has considerable permutation symmetry. For example: the start and end of a line segment could be swapped with no change to the resultant image; if the composition is non-sequential the order of the rendered primitives could be permuted; etc.}; see \cref{sup:optim-challenges} for more discussion. Additionally, the magnitude of gradients can become vanishingly small, which is particularly problematic with fixed-precision arithmetic; this problem can however be overcome as we demonstrate in the following sections.

\begin{table}[b]
    \centering
    \begin{tabular}{lccc}	
        \toprule
        $\mathcal{L}=$ 
        & $\mathrm{MSE}$ & $\mathrm{SSMSE}$ & $\mathrm{BlurMSE}$\\ \midrule
        $\mathcal{L}\left(\adjustbox{valign=c,margin=1mm}{\includegraphics[width=0.4cm]{images/halfhalf.tikz}}, \adjustbox{valign=c,margin=1mm}{\includegraphics[width=0.4cm]{images/greybox.tikz}}\right)$ & 0.25  & 0.42 & 0.13\\
         $\mathcal{L}\left(\adjustbox{valign=c,margin=1mm}{\includegraphics[width=0.4cm]{images/strips.tikz}}, \adjustbox{valign=c,margin=1mm}{\includegraphics[width=0.4cm]{images/greybox.tikz}}\right)$ & 0.25 & 0.25 & 0.02\\
         \bottomrule
    \end{tabular}
    \caption{Loss functions incorporating scale can overcome limitations of MSE and induce gradients.}
    \label{tab:losses}
\end{table}

\subsection{Loss Functions}
MSE loss is not the only possible choice; in fact, MSE has one significant disadvantage in that if we are drawing in black and white, but optimising a grey-level image, then the loss landscape would be very flat. This is illustrated in \cref{tab:losses} where it can be seen that both configurations of the generated image in the first input to the loss function produce exactly the same MSE value. Human vision does not suffer from the same problem; we can see that the two input images to the loss functions are clearly different. In addition, if we look from far enough away the striped example on the second row, the image and the target would begin to look the same to us. To build this phenomenon into the loss function we can incorporate a notion of spatial scale. We utilise two such approaches: BlurMSE, a single-scale version of MSE in which the input, and optionally the target are blurred by a Gaussian filter of a predetermined standard deviation; and the SSMSE, a scale-space variant in which a scale-space (or optionally a scale pyramid) is built for both the input and target, and the loss is accumulated over all levels. Our implementation of the scale-space follows \citet{Lowe:2004:DIF:993451.996342}, and constructs a space with octaves defined by a doubling of the standard deviation, and a fixed number of intervals per octave. As can be seen from \cref{tab:losses}, the losses incorporating scale produce smaller values for the striped input image on the second row, compared to the half-half image on the first row, thus indicating usable gradient information.

A Gaussian scale-space alone is not necessarily a good measure of how a human perceives an image as it fundamentally only helps capture areas of light and dark, and ensures they are shaded accordingly in the generated image. Many perceptually motivated distance metrics have been proposed in the past, such as the well-known SSIM~\citep{1284395} and its variants. More recently, it has been shown that features from deep convolutional networks can correlate well with human perceptual judgements of image similarity, and this has motivated the development of CNN-based perceptual losses like LPIPS~\citep{Zhang_2018_CVPR}. Because a loss based on deep features would inherently be differentiable, we can utilise it as an objective when optimising primitive parameters that define an image.

\begin{figure*}[ht]
    \centering
    \begin{minipage}[t]{0.49\textwidth}
    \begin{subfigure}[t]{.49\columnwidth}
        \centering
        \includegraphics[width=0.75\columnwidth]{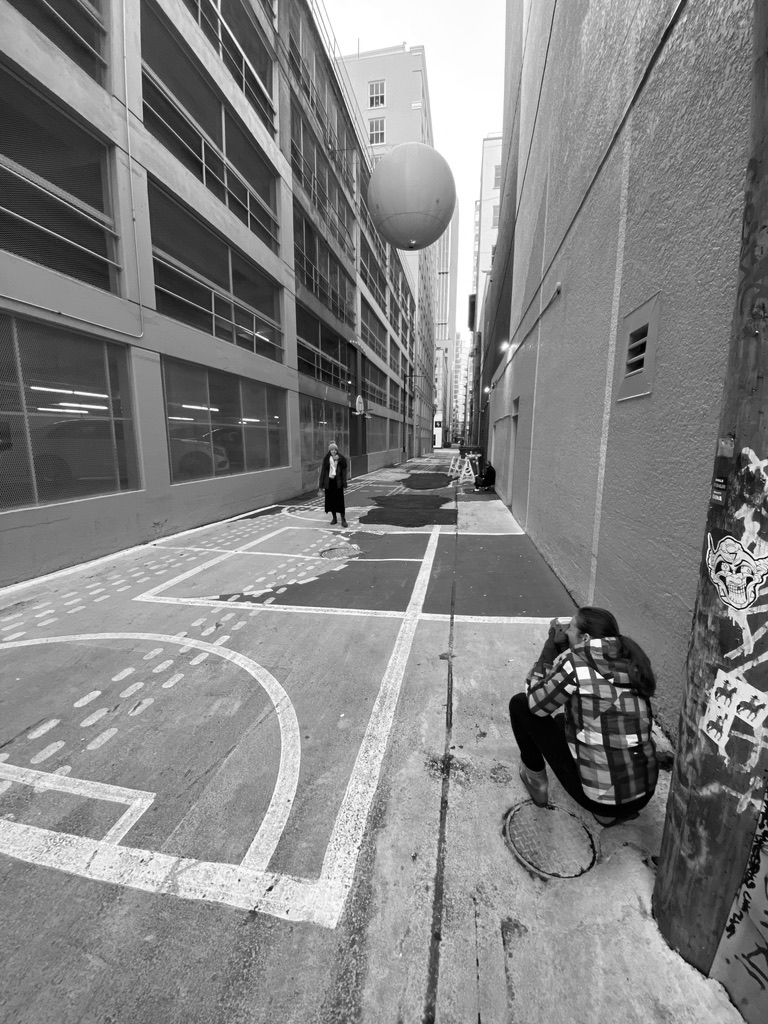}
    \caption{Photo}\label{fig:photo}
    \end{subfigure}\hfill%
    \begin{subfigure}[t]{.49\columnwidth}
        \centering
        \includegraphics[width=0.71\columnwidth]{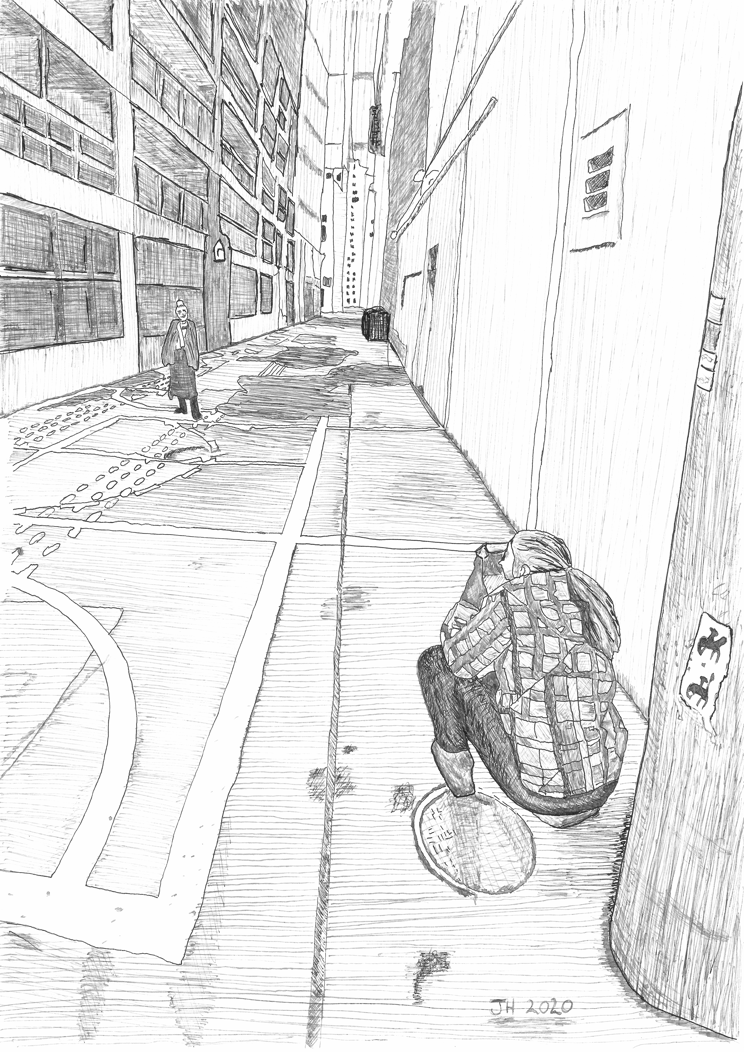}
        \caption{Pen \& Ink Sketch}\label{fig:sketch}
  \end{subfigure}
  \caption{Pictures of the author, by the other author.}\label{fig:sketches}
  \end{minipage}
  \begin{minipage}[t]{0.49\textwidth}
    \begin{subfigure}[t]{.49\columnwidth}
        \centering
        \includegraphics[width=0.75\columnwidth]{imageopt/points.pdf}
    \caption{1000 Points}\label{fig:points}
    \end{subfigure}\hfill%
    \begin{subfigure}[t]{.49\columnwidth}
        \centering
        \includegraphics[width=0.75\columnwidth]{imageopt/shortlines.pdf}
        \caption{1000 lines}\label{fig:shortlines}
  \end{subfigure}
  \caption{Optimising against \cref{fig:photo} using BlurMSE ($\sigma=1.0$).}\label{fig:blurredmseexamples}
  \vspace{0.5em}
  \end{minipage}
    \begin{subfigure}[b]{.19\textwidth}
        \centering
        \includegraphics[width=0.9\columnwidth]{imageopt/500lines-init.pdf}
        \caption{Initialisation}\label{fig:randinit}
    \end{subfigure}
    \begin{subfigure}[b]{.19\textwidth}
        \centering
        \includegraphics[width=0.9\columnwidth]{images/imageopt/500lines-mse.pdf}
        \caption{MSE}
    \end{subfigure}
    \begin{subfigure}[b]{.19\textwidth}
        \centering
        \includegraphics[width=0.9\columnwidth]{imageopt/500lines-blurredmse1.0ns.pdf}
        \caption{BlurMSE ($\sigma=1.0$)}
    \end{subfigure}
    \begin{subfigure}[b]{.19\textwidth}
        \centering
        \includegraphics[width=0.9\columnwidth]{images/imageopt/500lines-blurredmse3.0ns.pdf}
        \caption{BlurMSE ($\sigma=3.0$)}
    \end{subfigure}
    \begin{subfigure}[b]{.19\textwidth}
        \centering
        \includegraphics[width=0.9\columnwidth]{images/imageopt/500lines-blurredmse5.0ns.pdf}
        \caption{BlurMSE ($\sigma=5.0$)}
    \end{subfigure}
    
    \vspace{0.8em}%
    
    \begin{subfigure}[t]{.19\textwidth}
        \centering
        \includegraphics[width=0.9\columnwidth]{images/imageopt/500lines-pyrmseO5I1.pdf}
        \caption{SSMSE, 1i/o}
    \end{subfigure}
    \begin{subfigure}[t]{.19\textwidth}
        \centering
        \includegraphics[width=0.9\columnwidth]{images/imageopt/500lines-pyrmseO5I2.pdf}
        \caption{SSMSE, 2i/o}
    \end{subfigure}
    \begin{subfigure}[t]{.19\textwidth}
        \centering
        \includegraphics[width=0.9\columnwidth]{images/imageopt/500lines-pyrmseO5I4.pdf}
        \caption{SSMSE, 4i/o}
    \end{subfigure}
    \begin{subfigure}[t]{.19\textwidth}
        \centering
        \includegraphics[width=0.9\columnwidth]{images/imageopt/500lines-lpips-alex.pdf}
        \caption{LPIPS(AlexNet)}
    \end{subfigure}
    \begin{subfigure}[t]{.19\textwidth}
        \centering
        \includegraphics[width=0.9\columnwidth]{images/imageopt/500lines-lpips-vgg.pdf}
        \caption{LPIPS(VGG)}
    \end{subfigure}
  \caption{Images created by optimising parameters using gradient descent with different losses. Parameters optimised to fit the photo shown in \cref{fig:photo} starting from the random lines in \cref{fig:randinit}. All images using SSMSE use 5 octaves, and `i/o' abbreviates the number of intervals per octave. Note that regular MSE is just BlurMSE with $\sigma=0$.}\label{fig:directoptlosses}
\end{figure*}

\subsection{Examples: image based optimisation}
To demonstrate the effectiveness of our approach for optimising primitives against a real image we provide a number of examples. All of the generated images in \cref{fig:blurredmseexamples,fig:directoptlosses} utilise the $200\times266$ pixels input image in \cref{fig:photo} as the target image to optimise against. Points and pixels are optimised to have a 1-pixel diameter/thickness in the image space. The domain of the world-space is constrained to [-1,1] on the y-axis and scaled proportionally on the x-axis. All generated examples were optimised using Adam~\citep{DBLP:journals/corr/KingmaB14} with a learning rate of 0.01 for 500 iterations. \Cref{fig:blurredmseexamples} shows the results from optimising 1000 points and 1000 lines using blurred MSE loss and demonstrates the overall effect that can be achieved. \Cref{fig:directoptlosses} shows the effect of optimising 500 line segments from the same starting point using a range of different losses.
%

It is instructive to compare how the automatically generated sketches compare to an image drawn by a human. \Cref{fig:sketch} is a hand-drawn pen and ink sketch of the same scene as used in the generation of \cref{fig:blurredmseexamples,fig:directoptlosses}. It is clear that all of the sketches broadly capture the overall structure of the scene and areas of light and dark. However, there are significant differences in the way this is captured. The losses based on MSE (including scale-space and blurred) all display the same trait of capturing the local intensity, although this is much more pronounced in the scale-space and blur variants, which also capture more detail. Changing the number of intervals per octave in the scale-space losses has very little overall effect (subtle changes around the `balloon'). The perceptual loss using AlexNet captures highly local structure, but overall the resultant image is perhaps the least perceptually similar (or interpretable) of all the images. The perceptual loss using VGG captures a lot of the structure of the image; it is interesting how much of the broad shape information is captured, and how areas of light and dark are also represented. In addition, we can observe that the overall brightness on the right-hand side is lighter than the left, mimicking the human-drawn sketch, even though the raw grey-level values in the input image are similar on both sides. The differences between the two perceptual losses reflect the observation that the VGG variant is closer to traditional notions of perceptual difference when used for optimisation~\citep{Zhang_2018_CVPR}. Related to the observation that the VGG model seems to capture shape information rather well, we wonder if direct optimisation in the way we have performed it might lead to a new way to probe the (lack of) shape bias in different neural architectures~\citep{geirhos2018imagenettrained}. This 
could ultimately help us move closer to networks that robustly recognise objects from both sketches and photographs.

\section{Autotracing autoencoders}
\label{sec:ae}

We next look at models that learn to perform autotracing of handwritten characters with only self-supervision. The structure of our autotracing model, shown in \cref{fig:ae}, is similar to that of a standard autoencoder, with two main components: an image encoder that creates a latent encoding, and a parameter decoder that decodes a latent vector to `stroke data'. This stroke data is then rasterised into the output image. Both the encoder and parameter decoder have learnable parameters, but the rasterisation is entirely fixed.

We next demonstrate a series of decoders which allow for different approaches to drawing. For example, we consider stroke parametrisation functions such as independent straight lines/curves, connected lines/curves through a series of consecutive points, and sets of points with learned connections between them. These models lay the groundwork for future exploration of learned, differentiable models of sketching that are more similar to how humans write/draw that \eg address the challenges set out by \citet{lake2015human}.


\paragraph{Encoders.} For experiments on MNIST \citep{lecun1998mnist}, we present results using a simple multi-layer perceptron encoder network. For more complex characters of Omniglot~\citep{lake2015human}, a convolutional network is preferred. When comparing against StrokeNet~\citep{zheng2018strokenet} (\cref{Table:scaledMNISTcompar}), we replicate their VGG-like Encoder. Full model details are provided in \cref{sup:encoders}.

\begin{table*}[t]

\begin{subtable}{0.49\textwidth}
\centering

\begin{tabular}{lccccr}	
\toprule
Decoder & \#P & \#S & \#L & MSE & Acc.\\
\midrule
Line & 10 & 1 & 5 &0.0195 &94.06\% \\
PolyLine  & 16 & 15 & 1 & 0.0225 & 93.27\%\\
PolyConnect & 16 & - & - & 0.0118 &96.47\%\\
CRS & 16 & 14 & 1 & 0.0208 & 94.63\%\\
B\'ezier & 20 & 1 & 5 & 0.0136 & 96.34\% \\
B\'ezierConnect & 16 & - & - & 0.0116 & 96.43\%\\

\bottomrule
\end{tabular}
\caption{MNIST Test Dataset (baseline unencoded acc. 98.60\%).}
\label{Table:DecoderComparison}
\end{subtable}
\hfill%
\begin{subtable}{0.49\textwidth}
\centering
\begin{tabular}{lccccr}	
\toprule
Model & Steps &\#P & \#S &\#L & Acc.\\
\midrule
StrokeNet \citep{zheng2018strokenet} & 3 (SN) & 16 & 14 & 1   & 95.25\% \\
StrokeNet \citep{zheng2018strokenet} & 1 & 16 & 14 & 1   & 97.75\% \\
Ours, CRS & 1 & 16 & 14 & 1          & 97.12\% \\
Ours, B\'ezier & 3 (GRU) & 4 & 1 & 1   & 96.97\%\\ 
Ours, B\'ezier & 1 & 7 & 2 & 2         & 98.28\%\\ 
Ours, B\'ezier & 1 & 43 & 14 & 1       & 97.94\%\\ 
\bottomrule
\end{tabular}

\caption{Scaled MNIST Dataset (baseline unencoded acc. 98.58\%).}
\label{Table:scaledMNISTcompar}
\end{subtable}

\caption{Reconstruction performance of parameterisations, measured by MSE and classification accuracy with a classifier trained on unencoded training sets of the respective datasets. \#* indicates the number of (L)ines, (S)egments, and (P)oints. 
All Scaled MNIST models use the same `StrokeNet Agent' architecture~\citep{zheng2018strokenet} to map images to primitive parameters.}

\end{table*}

\paragraph{Decoders.}
Our decoder networks allow for different parametrisations of `stroke data' that is then used by the rasteriser described in \cref{sec:relaxedrast}. The decoder transforms a vector encoding of the input image to lists of stroke primitives which aim to reproduce the input image when rasterised. 
In the simplest case, the latent vector can be decoded to a fixed number of line segments (\textit{LineDecoder}), each defined by their start and end points. Next, we provide \textit{PolyLineDecoder}, for which a stroke is represented as a sequence of consecutive points.  Instead of line segments, we can choose to use curves parameterised as Catmull-Rom splines (\textit{CRSDecoder}) or B\'ezier curves (\textit{B\'ezierDecoder}). In both cases, we can control the number of joined curve segments by specifying how many points (CRS) or segments (B\'ezier) are used. To allow more flexibility in modelling, we have also explored decoders which incorporate sub-networks to learn to produce a set of 2d points, and the upper-triangular portion of a soft connection matrix between points (optionally including the diagonal). The network producing the connection matrix uses a sigmoid to ensure values are between 0 and 1. To utilise the connection matrix, all possible combinations of lines are rasterised and are multiplied by the appropriate connection weight before composition (\textit{PolyConnect}). In the case of B\'ezier curves (\textit{B\'ezierConnect}) each point in the connection matrix corresponds to both an end point and its corresponding control point, and when drawing curves, the end point is drawn using the mirror of its control point allowing for smooth multiple-segment curves to be created. \citet{zheng2018strokenet} proposed a recurrent model using a visual working memory; the network is presented at each timestep with the features of the target image, together with the current canvas, which is then encoded, concatenated with the input, and transformed to the parameters of a new stroke which is rendered and overlaid on the canvas. We experimented with this approach but found it hard to train and computationally expensive, so we also investigated a simple GRU~\citep{cho-etal-2014-learning} based RNN which is fed a target image's encoding as its initial hidden state, along with a projection of a zeroed input. The GRU output is projected to a set of B\'ezier curve parameters for rendering, and also re-projected for input at the next time step. Full details are given in \cref{sup:decoders}. 

Table~\ref{Table:DecoderComparison} shows the effect of different stroke parametrisations on MNIST (reconstructions shown in \cref{sup:MNISTrecon}). As an objective measure, we compute the classification accuracy of rasterised sketches from the test set using a classifier (baseline accuracy of 98.6\%); reconstructions that capture the character should have higher accuracy. \textit{Connect} models, which generate strokes based on a learned connection matrix for the given number of points,  perform best due to the flexibility of deciding which points should be joined in a line/curve segment.  Following \citet{zheng2018strokenet} we perform a similar experiment on their scaled MNIST dataset (see \cref{sup:strokenet}), and also show results using the pretrained StrokeNet models that are publicly available. The accuracies of all models are high indicating good reconstructions, but we note that MNIST doesn't require complex decoders. 

\begin{figure}[b]
    \centering
    \begin{subfigure}[t]{.25\columnwidth}
        \centering
        \includegraphics[height=2.1cm]{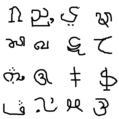}
        \caption{Val}
        \label{fig:valsamples}
    \end{subfigure}\hfill%
    \begin{subfigure}[t]{.25\columnwidth}
        \centering
        \includegraphics[height=2.1cm]{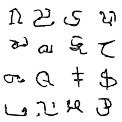}
        \caption{Val rec.}
        \label{fig:valrecon}
    \end{subfigure}\hfill%
    \begin{subfigure}[t]{.25\columnwidth}
        \centering
        \includegraphics[height=2.1cm]{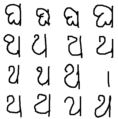}
        \caption{Test}
        \label{fig:testsamples}
    \end{subfigure}\hfill%
    \begin{subfigure}[t]{.25\columnwidth}
        \centering
        \includegraphics[height=2.1cm]{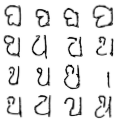}
        \caption{Test rec.}
        \label{fig:testrecon}
    \end{subfigure}\hfill%
    
  \caption{28 pixel Omniglot validation and test data samples and \textit{B\'ezier} model (3 segment, 5 line) reconstructions.}
  \label{fig:omniglotsketches}
\end{figure}

Figure~\ref{fig:omniglotsketches} illustrates reconstruction of Omniglot~\citep{lake2015human}. Note that the test set contains alphabets completely disjoint from training/validation. Some small details of the characters are missing, and it is clear that the models do not always choose to draw stokes in the way a human would, but the performance is generally good (see \cref{sup:omniglotresults}). Similar analysis on additional datasets is shown in \cref{sup:additional-AE-results}.

\section{Future Directions}
\label{sec:concl}

We have presented a derivation of a bottom-up differentiable approach to rasterising vector primitives into images, that allows gradients to flow through every pixel in the image to the underlying primitive's parameters. Our approach allows us to construct end-to-end models of vision that learn primitive parameters directly from raster images. Further, we have demonstrated how effective sketch generation can be achieved with different losses, and how parameterisations can change what a model learns.

Our approach is only a building block towards future applications and research. Our own motivation for designing this approach is to use it to explore writing and visual communication, although there are undoubtedly many potential use-cases.  For us, questions to be answered next involve looking at how we might build models that can learn to produce the appropriate number of strokes (and choose between different types of primitive). As part of this, it is clear that reconstruction performance alone should not be the key driver of gradient; the ability to communicate information is more important. Both attention and weak supervision to better mimic humans are also key to this endeavour.

{\small
\bibliographystyle{ieee_fullname}
\bibliography{main}
}
\flushcolsend
\newpage

\appendix   
\counterwithin{equation}{section}
\counterwithin{table}{section}
\renewcommand\thesection{\Alph{section}}
\renewcommand{\thefigure}{\Roman{figure}}

\newcommand{\makeextravisrow}[4]{
\raisebox{-.5\height}{\rule{0pt}{2.8cm}}\rotatebox[origin=c]{90}{#4} &%
\raisebox{-.5\height}{\includegraphics[width=0.2\textwidth]{images/imageopt-extra/extras/#1/#2-500#3-fixed-bw/final.pdf}} &%
\raisebox{-.5\height}{\includegraphics[width=0.2\textwidth]{images/imageopt-extra/extras/#1/#2-500#3-fixed-col/final.pdf}} &%
\raisebox{-.5\height}{\includegraphics[width=0.19\textwidth]{images/imageopt-extra/extras/#1/#2-500#3-learned-bw/final.pdf}} &%
\raisebox{-.5\height}{\includegraphics[width=0.2\textwidth]{images/imageopt-extra/extras/#1/#2-500#3-learned-col/final.pdf}} \\	
}	
\newcounter{extrascounter}	
\newcommand{\makeextravis}[1]{	
\begin{figure*}[ph]	
\centering	
\begin{tabular}{cccccc}	
& & \multicolumn{4}{c}{\includegraphics[width=4.1cm]{images/imageopt-extra/extras/#1.JPG}} \\[1em]	
\toprule	
& & \multicolumn{2}{c}{Fixed Width} & \multicolumn{2}{c}{Learned Width}  \\	
& & Black \& White & Uniform Colour & Black \& White & Uniform Colour  \\	
\midrule	
\multirow{14.5}{*}{\rotatebox[origin=c]{90}{MSE}} & \makeextravisrow{#1}{mse}{pts}{500 Points} 	
& \makeextravisrow{#1}{mse}{lines}{500 Lines}	
& \makeextravisrow{#1}{mse}{crs}{500 Curves}	
\midrule	
\multirow{14.5}{*}{\rotatebox[origin=c]{90}{LPIPS(VGG)}} &	
\makeextravisrow{#1}{lpipsvgg}{pts}{500 Points}	
& \makeextravisrow{#1}{lpipsvgg}{lines}{500 Lines}	
& \makeextravisrow{#1}{lpipsvgg}{crs}{500 Curves}	
\bottomrule	
\end{tabular}%
\refstepcounter{extrascounter}	
\caption{Examples of image optimisation (\roman{extrascounter}).}	
\label{fig:extra#1}	
\end{figure*}	
}

\twocolumn[{%
\begin{center}
    {\Large \bf Appendices \par}
    \vspace*{24pt}
\end{center}
\begin{center}
    \captionsetup{type=figure}
    \includegraphics[width=0.8\textwidth]{images/imageopt/bathers.pdf}
    \caption{Is this what Georges might choose to paint if he decided to paint with straight lines rather than points? This image was created by taking a photo of the original `Une baignade à Asnières' from Wikipedia (\url{https://en.wikipedia.org/wiki/Bathers_at_Asnieres}) and optimising 500 uniformly coloured lines to fit the image.}
    \label{fig:bathers-big}
\end{center}
}]

\section{Relating $\sigma$ to line thickness.} \label{sup:thickness}

Consider the 1D rasterisation of a point of size $t$ given by the scaled unit box function $\Pi(x/t)$ and the relaxed rasterisation given by $exp(-d^2(x)/\sigma^2)$ as illustrated in \cref{fig:linewidth}.
\begin{figure}[b]
    \centering
    \begin{tikzpicture}
    \begin{axis}[ 
    xlabel=$x$,
    xmin=-7,
    xmax=7,
    samples=200,
    width=\columnwidth,
    height=100,
    grid=major,
    ymajorgrids=false,
    xtick distance=1,
    grid style={line width=.1pt, draw=gray!10},
    major grid style={line width=.2pt,draw=gray!50},
    declare function={
    box(\x)= and(\x>=-0.5, \x<=0.5);
    },
    legend style={at={(0.5,-0.5)}, anchor=north} 
  ] 
    \addplot[domain=-7:7, dashed] {e^(- x^2 / 1)}; 
    \addplot[domain=-7:7, dotted] {e^(- x^2 / 4)}; 
    \addplot[domain=-7:7, solid] {box(x / 2)}; 
    
    \legend{$\exp(-x^2/\sigma^2); \sigma^2=1$, $\exp(-x^2/\sigma^2); \sigma^2=4$, $\Pi(x/t); t=2$} 
  \end{axis}
    \end{tikzpicture}
    \caption{Illustration of a target point of thickness $t=2$ pixels and approximations with the $exp$ rasterisation function with different $\sigma^2$ values.}
    \label{fig:linewidth}
\end{figure}
\noindent We want to find a relationship between the value of $t$ and $\sigma^2$ when trying to minimise the squared difference of the functions across the entire domain $x$,
\begin{equation}
    \begin{aligned}
    \min_{\sigma^2} \quad & \int_{-\infty}^{\infty} (e^{-x^2 / \sigma^2} - \Pi(x/t))^2\,dx\\
    \textrm{s.t.} \quad & t > 0 \\
                        & \sigma^2 > 0 \;.
    \end{aligned}
\end{equation}
The integral term can be expanded and evaluated as follows (assuming the constraints $t>0$ and $\sigma^2>0$):
\begin{align}
    & \int_{-\infty}^{\infty} (e^{-x^2 / \sigma^2} - \Pi(x/t))^2\,dx \nonumber \\
    &= \int_{-\infty}^{\infty} e^{-2x^2 / \sigma^2}
    -2 \Pi(x/t) e^{-x^2 / \sigma^2}
    + \Pi(x/t)^2\,dx \nonumber \\
    &= \sigma\sqrt{\frac{\pi}{2}} - 2 \sigma \sqrt{\pi} \erf\left(\frac{t}{2\sigma}\right) + t \;.
\end{align}
Now, differentiating and setting to zero gives
\begin{align}
    0 &= \frac{d\left(\sigma\sqrt{\frac{\pi}{2}} - 2 \sigma \sqrt{\pi} \erf\left(\frac{t}{2\sigma}\right) + t\right)}{d\sigma} \nonumber \\
    &= \sqrt{\frac{\pi}{2}} - 2 \sqrt{\pi} \frac{d\left(\sigma \erf\left(\frac{t}{2\sigma}\right)\right)}{d\sigma} \nonumber \\
    &= \sqrt{\frac{\pi}{2}} - 2 \sqrt{\pi} \left( \erf\left(\frac{t}{2\sigma}\right) + \sigma \frac{d\left(\erf\left(\frac{t}{2\sigma}\right)\right)}{d\sigma} \right) \nonumber \\
    &= \sqrt{\frac{\pi}{2}} - 2 \sqrt{\pi} \erf\left(\frac{t}{2\sigma}\right) - 2 \sqrt{\pi} \sigma \left( - \frac{ t e^{-t^2/(4\sigma^2)}}{\sigma^2\sqrt{\pi}}\right) \nonumber \\
    &= \sqrt{\frac{\pi}{2}} - 2 \sqrt{\pi} \erf\left(\frac{t}{2\sigma}\right) + \frac{ 2 t e^{-t^2/(4\sigma^2)}}{\sigma}\;. \label{eqn:thickderiv}
\end{align}
Noting the common factors of $t/\sigma$ in \cref{eqn:thickderiv} we can write the right hand side as an expression in terms of $c=t/\sigma$:
\begin{equation}
    \sqrt{\frac{\pi}{2}} - 2 \sqrt{\pi} \erf\left(\frac{c}{2}\right) + 2 c e^{-c^2/4} \;. \label{eqn:thickderiv2}
\end{equation}
As shown in \cref{fig:linethicknessratio} this expression is monotonically decreasing and has a single root, which can be estimated numerically as $c \approx 1.820657$. 

\begin{figure}[H]
    \centering
    \begin{tikzpicture}
    \begin{axis}[ 
    xlabel=$c$,
    xmin=-15,
    xmax=15,
    samples=200,
    width=\columnwidth,
    height=100,
    grid=major,
    ymajorgrids=true,
    grid style={line width=.1pt, draw=gray!10},
    major grid style={line width=.2pt,draw=gray!50},
    declare function={erf(\x)=%
      (1+(e^(-(\x*\x))*(-265.057+abs(\x)*(-135.065+abs(\x)%
      *(-59.646+(-6.84727-0.777889*abs(\x))*abs(\x)))))%
      /(3.05259+abs(\x))^5)*(\x>0?1:-1);
    }
  ] 
    \addplot[domain=-15:15, solid] {(pi/2)^0.5 + 2 * x * e^(- x^2 / 4) - 2 * pi^0.5 * erf(x/2)}; 
  \end{axis}
    \end{tikzpicture}
    \caption{Plot of \cref{eqn:thickderiv2}.}
    \label{fig:linethicknessratio}
\end{figure}
\noindent This implies the relationship between $t$ and $\sigma$ is linear: $\sigma \approx 0.54925 t\,\forall\, t>0$. This can be easily verified by substituting $\sigma = 0.54925 t$ in \cref{eqn:thickderiv}.

\section{Curve Parameterisations}
\label{sup:curveparam}
Curves are often represented mathematically by parametric functions $C(t)$ that give the coordinates of the curve for values of $t$, commonly in the closed interval $[0,1]$, with $t=0$ representing the start point and $t=1$ representing the end point of the curve. Curves are often parameterised as the coefficients of polynomial basis functions, either in Hermite (e.g. the curve is a linear combination of Hermite bases) or B\'ezier form (the curve is represented as a linear combination of Bernstein bases).  The following parametric curve formulations are commonly used in computer graphics (and can all be used in our differentiable rasterisation approach):

\paragraph{Quadratic B\'ezier Curves} are parameterised by three points: the start of the curve $\bm{P}_0$, the end of the curve $\bm{P}_2$ and the control point $\bm{P}_1$ which is the point the tangents to the curve at $\bm{P}_0$ and $\bm{P}_2$ intersect. The curve would not normally pass through $\bm{P}_1$. The curve can be thought of as leaving $\bm{P}_0$ in the direction of $\bm{P}_1$ and gradually bending to arrive at $\bm{P}_2$ from the direction of $\bm{P}_1$. The quadratic B\'ezier is defined as:
\begin{align}
\operatorname{C_{bez^2}}(t, \bm\theta) =& (1 - t)^{2}\bm{P}_0 + 2(1 - t)t\bm{P}_1 + t^{2}\bm{P}_2 \\
\shortintertext{where}
\bm{\theta} =& \left[\bm{P}_0 | \bm{P}_1 | \bm{P}_2 \right].\nonumber
\end{align}
    
\paragraph{Cubic B\'ezier Curves} are defined by four points: $\bm{P}_0$ is the start of the curve; $\bm{P}_1$ is the first control point and indicates the direction the curve leaves $\bm{P}_0$ from; $\bm{P}_2$ is the second control point and indicates the direction that the curve arrives at the final end point $\bm{P}_3$ from. The curve would not normally pass through either control point. The cubic B\'ezier is defined as:
\begin{align}
\operatorname{C_{bez^3}}(t, \bm\theta) =& (1-t)^3\bm{P}_0+3(1-t)^2t\bm{P}_1 \nonumber \\
&\;\; +3(1-t)t^2\bm{P}_2+t^3\bm{P}_3 \\
\shortintertext{where}
\bm{\theta} =& \left[\bm{P}_0 | \bm{P}_1 | \bm{P}_2 | \bm{P}_3\right].\nonumber
\end{align}
    
\paragraph{Catmull-Rom Splines} parameterise a curve by 4 points which the curve passes through smoothly. The curve is only drawn between the middle pair of points:
    \begin{align}
\operatorname{C_{crs}}(t,\bm\theta) &= \frac{t_{2}-t}{t_{2}-t_1}\bm{B}_1+\frac{t-t_1}{t_{2}-t_1}\bm{B}_2\\
\shortintertext{where}
\bm{B}_1 &= \frac{t_{2}-t}{t_{2}-t_0}\bm{A}_1+\frac{t-t_0}{t_{2}-t_0}\bm{A}_2 \nonumber\\
\bm{B}_2 &= \frac{t_{3}-t}{t_{3}-t_1}\bm{A}_2+\frac{t-t_1}{t_{3}-t_1}\bm{A}_3 \nonumber\\
\bm{A}_1 &= \frac{t_{1}-t}{t_{1}-t_0}\bm{P}_0+\frac{t-t_0}{t_{1}-t_0}\bm{P}_1 \nonumber\\
\bm{A}_2 &= \frac{t_{2}-t}{t_{2}-t_1}\bm{P}_1+\frac{t-t_1}{t_{2}-t_1}\bm{P}_2 \nonumber\\
\bm{A}_3 &= \frac{t_{3}-t}{t_{3}-t_2}\bm{P}_2+\frac{t-t_2}{t_{3}-t_2}\bm{P}_3 \nonumber\\
t_{0} &= 0 \nonumber\\
t_{i+1} &= || \bm{P}_{i+1} - \bm{P}_i ||_2^{\alpha} + t_i \nonumber\\
\shortintertext{and}
\bm{\theta}&=\left[\bm{P}_0 | \bm{P}_1 | \bm{P}_2 | \bm{P}_3\right].\nonumber
    \end{align}
The \textit{centripetal} Catmull-Rom spline sets $\alpha$ to $0.5$, which has the advantage that cusps or self-intersections cannot be formed in the curve.

\begin{algorithm*}[t]
\SetKwFor{For}{for (}{) $\lbrace$}{$\rbrace$}
\SetNoFillComment 
\Fn{\FCurveDistPolyline{$\bm\theta$, $C$, $\bm{n}$, segments}}{
  \KwData{\\
  \addtolength{\leftskip}{1em}
    $\bm\theta$: curve parameters.\\
    $C$: function defining coordinates of curve at a distance $0 \leq t \leq 1$ along it.\\ 
    $\bm{n}$: coordinate to compute distance from. \\
    \ArgSty{segments}: number of line segments to use in the approximation. \\
  }
  \KwResult{the square of the minimum distance between $\bm{n}$ and the curve.}
  \BlankLine
  \ArgSty{mindist} $\gets \infty$
  
  \For{$i = 1;\ i \leq$ \ArgSty{segments} $;\ i = i + 1$}{
    $t_0 \gets (i - 1)\;/ $ \ArgSty{segments} \\
    $t_1 \gets (i)\;/ $ \ArgSty{segments} \\
    
    \ArgSty{dist} $\gets \mindisttolineseg(\bm{n}, C(t_0, \bm\theta), C(t_1, \bm\theta))$ \tcp{See \cref{eqn:closestPointLine}}
    \If{dist $<$ mindist}{
      \ArgSty{mindist} $\gets$ \ArgSty{dist} \\
    }
  }
  \BlankLine
  \KwRet{mindist}
}
 \caption{\label{alg:polyline}Polyline approximation for the closest point on a curve. This approximation breaks the curve into \textit{segments} uniform-$\Delta t$ line segments, and might be sub-optimal in areas of high curvature (if such areas were to exist, then an adaptive variant of this algorithm could instead be used).}
\end{algorithm*}

\section{Computing the Squared Euclidean Distance Transform for a curve}
\label{sup:curveedt}

In general, it is not possible to write a closed form expression for the (squared) distance of an arbitrary point, $\bm n$ to the closest point on a curve, $C(t)$, 
\begin{equation}
    \begin{aligned}
        \mindisttocurve(\bm{n}) = \min_{t} \quad & ||C(t) - \bm{n}||^2_2 \\
        \textrm{s.t.} \quad & 0 \leq t \leq 1 \; .\\
    \end{aligned}
\end{equation}
Potential approaches to computing this would for example be through a polyline approximation (see \cref{alg:polyline}), a recursive brute force search (see \cref{alg:curvebrute}) or a method based on finding the roots of the polynomial given by the derivative
\begin{equation}
    \frac{d}{dt} ||C(t) - \bm{n}||^2_2 \;.
    \label{eqn:curvederiv}
\end{equation}

In the latter case, the root-finding itself could be achieved in several ways; for example, by computing the real eigenvalues of the companion matrix formed from \cref{eqn:curvederiv} that lie between $0$ and $1$ and selecting the one that gives minimum distance, or by locating two values of $t$ that give opposing signs of \cref{eqn:curvederiv} and applying the bisection method. Another potential alternative is the method proposed by \citet{10.1145/3414685.3417871} which uses bisection with the Newton-Raphson method, with the initial guess computed using isolator polynomials~\citep{Sederberg:1994:IP}.

The challenge of all the latter approaches is efficient vectorised batch implementation, whereby computation of distance transforms (the computation of the minimum distance to a curve for all points in the image space) is performed for a \textit{batch} of curves in parallel making efficient use of many-core hardware. An approach based on root finding using the real eigenvalues of the companion matrix, for example, should ultimately prove to be more accurate than a polyline approximation, and potentially better and faster than the brute-force search, however at the time of writing there are not any hardware optimised batch generalised Eigendecomposition (GEVD) implementations available; a batch GEVD  implementation (for small matrices) is necessary as the decomposition would have to be computed for every pixel~$\bm n \in \mathcal{I}$.

Currently, we have proof-of-concept implementations using the former polyline approximation and brute force approaches, and these are both vectorised to run on many-core (particularly GPU) hardware. The polyline approximation is in general faster (obviously both the polyline and brute force approaches allow the degree of precision to be adjusted, and that changes the computational complexity), but it does have a potential disadvantage that the approximation can introduce degeneracies whereby a small change in a curve's parameters cause a topological change in the polyline approximation. In practice, however, we have not found this to be a problem in all our experiments with handwritten characters, which all use a 10-segment polyline approximation for each curve segment that is drawn.

\begin{algorithm*}[t]
\Fn{\FCurveDistRecurs{$\bm\theta$, $C$, $\bm{n}$, $t_{min}$, $t_{max}$, iters, slices, mindist=$\infty$}}{
  \KwData{\\
  \addtolength{\leftskip}{1em}
    $\bm\theta$: curve parameters.\\
    $C$: function defining coordinates of curve at a distance $0 \leq t \leq 1$ along it.\\ 
    $\bm{n}$: coordinate to compute distance from. \\
    $t_{min}$: starting value of $t$ for the search. \\
    $t_{max}$: ending value of $t$ for the search. \\
    \ArgSty{iters}: number of iterations to perform. \\
    \ArgSty{slices}: number of intervals between $t_{min}$ and $t_{max}$  to compute the distance at. \\
    \ArgSty{mindist}: current minimum distance estimate. \\
  }
  \KwResult{the square of the minimum distance between $\bm{n}$ and the curve.}
  \BlankLine
  \If{iters $\leq 0$}{
    \KwRet{mindist}
    }
  \BlankLine
  $\Delta_t \gets (t_{max} - t_{min})\;/$ \ArgSty{slices} \\
  $t \gets t_{min}$ \\
  $t_{best} \gets t_{min}$ \\
  \BlankLine
  \Repeat{$t \geq t_{max}$}{
  \ArgSty{dist} $\gets ||C(t, \bm\theta) - \bm{n}||_2^2$ \\
  \If{dist $<$ mindist}{
    \ArgSty{mindist} $\gets$ \ArgSty{dist} \\
    $t_{best} \gets t$ \\
  }
  $t \gets t + \Delta_t$\\
  }
  \BlankLine
  \KwRet{\FCurveDistRecurs{$\bm\theta$, $C$, $\bm{n}$, $t_{best}-\Delta_t$, $t_{best}+\Delta_t$, iters$-1$, slices, mindist}}
}

 \caption{\label{alg:curvebrute}Recursive brute-force search for the closest point on a curve. This is approximate in the sense that if \emph{slices} is too small the wrong minima might be located, and that \emph{iters} controls the precision of the solution that is found.}
\end{algorithm*}

\section{Composition functions}
\label{app:compfcns}

The \emph{soft-or} composition operator (\cref{eqn:softor}) defined in \cref{ssec:compositions} provides a good model of drawing with an instrument like a black ink pen, where overlapping strokes are not visible. Such a function is invariant to the order of the strokes. We might however consider alternative drawing functions that enable different effects and models of drawing and blending, to be achieved. Here we discuss a few potential options, including the \emph{over} operator used for our colour drawing examples. Note the focus here is on drawing opaque \textit{colours}; compositions for colour with transparency are discussed in \cref{sec:coltransp}.

\subsection{The \textit{over} composition operator}
The first potential alternative approach to the soft-or would be to define a composition that respects the ordering of the images and `paints' each stroke \textit{over} the top of the other (whilst not allowing background 0 pixels to cover already filled pixels) from the background to the foreground. Taking inspiration from \citet{10.1145/800031.808606}'s methods for alpha composition of computer graphics we could define a composition of image $\bm{A}$ painted over image $\bm{B}$ as:
\begin{equation}
    \operatorname{c_{over}}(\bm{A}, \bm{B}) = \bm{A} + \bm{B}(\bm{1}-\bm{A}) \;.
\end{equation}
This function could then be applied recursively over a sequence of depth-ordered rasterisations to compose in the desired way:
\begin{equation}
    \operatorname{c_{over}}(\bm{I}_4, \operatorname{c_{over}}(\bm{I}_3, \operatorname{c_{over}}(\bm{I}^{(2)}, \bm{I}^{(1)})))  \;.
    \label{eqn:over-recur}
\end{equation}
This type of approach does however have a significant problem in terms of implementation: because it is recursive and sequential, it is not easily vectorised and introduces a significant processing bottleneck which makes it intractable to use with large numbers of images. This problem can be circumvented by rewriting\footnote{This was first noted by \citet{10.1145/1507149.1507160} for \citeauthor{10.1145/800031.808606}'s \textit{over} operator with an alpha channel (see also \cref{sec:coltransp}).} \cref{eqn:over-recur} as follows,
\begin{equation}
    \operatorname{c_{over}}(\bm{I}^{(1)}, \dots, \bm{I}^{(n)}) = \sum_{i=1}^n \bm{I}^{(i)} \odot \prod_{j=1}^{i-1} (\bm{1} - \bm{I}^{(j)}) \;.
    \label{eqn:over-unrolled}
\end{equation}
In this form, we can see that in essence the computation required consists of the calculation of the cumulative product of a difference, a multiplication, and a summation; all of which can be efficiently vectorised. For numerical stability, the cumulative product can be computed as the exponentiated sum of the log differences,
\begin{equation}
    \operatorname{c_{over}}(\dots) = \sum_{i=1}^n \bm{I}^{(i)} \odot \exp\left(\sum_{j=1}^{i-1} \log(\bm{1} - \bm{I}^{(j)}) \right) \;.
    \label{eqn:over-unrolled-log}
\end{equation}
The inner summation can easily be implemented using the \verb|cumsum| operator built into most tensor processing libraries; note, however, that standard implementations will likely include cumulative sum up to and including the $i$-th image, so this must then be subtracted to give the required value. Additional care must also be taken to avoid taking the logarithm of zero; in practice adding a small epsilon value suffices.

\subsection{The \textit{max} composition operator}
Another possible alternative composition would be to take the per-pixel maximum over the set of images:
\begin{align}
    &\operatorname{c_{max_{i,j}}}(\bm{I}^{(1)}, \bm{I}^{(2)}, \dots, \bm{I}^{(n)}) \nonumber \\
    & \;\;\;\; = \max(\bm{I}^{(1)}_{i,j}, \bm{I}^{(2)}_{i,j}, \dots, \bm{I}^{(n)}_{i,j}) \;.
\end{align}
Clearly this does not have usable gradients because of the $\max$, however, a suitable differentiable relaxation exists with the $\smoothmax$ function,
\begin{equation}
    \smoothmax(\bm{x}) = \softmax(\bm{x} / \tau)^\top \bm{x} \;,
    \label{eqn:smoothmax}
\end{equation}
where $\bm{x}$ is a vector of values to find the maximum of, and $\tau$ is a temperature parameter. As $\tau \to 0$, $ \smoothmax(\bm{x}) \to \max(\bm{x})$. \Cref{eqn:smoothmax} can be applied pixel-wise over a vector formed from the stacking of $[\bm{I}^{(1)}_{i,j},\bm{I}^{(2)}_{i,j},\dots,\bm{I}^{(n)}_{i,j}]$ to form a composition function:
\begin{align}
    &\operatorname{c_{smoothmax_{i,j}}}(\bm{I}^{(1)}, \bm{I}^{(2)}, \dots, \bm{I}^{(n)}) \nonumber\\
    &\;\;\;\; = \smoothmax([\bm{I}^{(1)}_{i,j},\bm{I}^{(2)}_{i,j},\dots,\bm{I}^{(n)}_{i,j}]^\top) \;.
\end{align}

\section{Extended Drawing}
\label{sup:extendrelax}

The main body of this paper focused on the act of drawing strokes in a differentiable manner and did not explore extensions to the model that would allow for more nuanced drawing --- for example in colour, and with different types of stroke. We demonstrate here how the framework we have already described can be extended to allow for more control over the drawings that are produced.

\subsection{Stroke width}
\Cref{sup:thickness} demonstrates that there is a direct relationship between stroke thickness and the $\sigma$ parameter used by the rasterisation function. In all the experimental results shown, we used the same fixed $\sigma$ for all strokes, although it should be immediately evident that this isn't a requirement, and that different strokes could have different $\sigma$ values, and thus different thicknesses.

Going further, the $\sigma$ value doesn't have to be a hyperparameter of the model; without changing anything within the rasterisation approach it is evident that one can compute gradients with respect to $\sigma$ for every stroke that is drawn. As such, it is entirely possible to learn the line thickness of each stroke (either independently or together) by appropriately parameterising the model. 

Real drawings sometimes exhibit a variation in stroke width along the length of a stroke; often this is a result of variations in pressure on the drawing instrument. It is possible to incorporate such variation into our drawing model by noting that our functions for both line segments and curves have a parameter $0 \leq t \leq 1$ along their length that can be used as an input to a function that produces different values of $\sigma$ along the length of the line (or equivalently we can modify the distance map). Such a function could be parameterised by \eg a simple neural network, and thus learned during the training or optimisation of a model.

\subsection{Colour}
\label{sec:colour}

Different shades of grey for individual strokes can be achieved by scalar multiplication of each stroke's raster with a grey-value before composition (note that soft-or would no longer necessarily be appropriate, so a different composition would likely be used). The grey-value could be learned or be a hyperparameter. 

To rasterise full-colour strokes, the simplest approach is to replicate the image for a rasterised stroke three times in the channel dimension, and then multiply by a tuple of values corresponding to the desired red, green, and blue values. Again, the parameters can be learned, as is illustrated in \cref{fig:bathers-big}, which uses the \textit{over} composition operation (\cref{eqn:over-unrolled-log}).

If we want to rasterise lines along which the colour changes, we can follow the same methodology for changing stroke width and learn functions that emit colour as a function of the relative position, $t$, along the stroke.

\subsection{Incorporating Transparency}
\label{sec:coltransp}

The differentiable rasterisation approach for colour described above can also be extended to deal with transparency. If we assume a pre-multiplied alpha colour model, where the the red, green and blue values of a pixel represent emission, and the alpha value represents occlusion, then we can directly use \citet{10.1145/800031.808606}'s compositing arithmetic. For example, the over operator with alpha,
\begin{align}
    c_o &= c_a + c_b(1 - \alpha_a) \nonumber \\
    \alpha_o &= \alpha_a + \alpha_b(1 - \alpha_a) \;,
\end{align}
allows for models that can learn appropriate colour and transparency for each stroke drawn.

	\section{Additional Optimisation Results}	
\label{app:extraopt}	
The extended drawing operators described in \cref{sup:extendrelax} can be utilised directly in the optimisation approach described and illustrated in \cref{sec:optimisation}. 	
\paragraph{Optimising Photographs.}	
In \cref{fig:extra1,fig:extra2,fig:extra3,fig:extra4,fig:extra5,fig:extra6,fig:extra8,fig:extra9,fig:extra10,fig:extra11,fig:extra12,fig:extra13,fig:extra15,fig:extra17} we present additional results on a wide variety of images showing the result of performing image optimisation with both the MSE loss and LPIPS(VGG) loss, together with different drawing configurations. There is not much additional to note from these examples than was already covered in \cref{sec:optimisation}, however we reiterate that the LPIPS(VGG) loss is strikingly good at giving results that capture strong perceptual features.	
\paragraph{Optimising Cartoons.}	
\textit{Optimising cartoons} is not well suited for the `direct optimisation' setup described in \cref{sec:optimisation}. Whilst the rasteriser can be used for this task, there are considerable inductive biases that would be useful to incorporate to make optimisation easier. For example, incorporating strong priors for initialisation and using a more sensible loss (probably one based on chamfer distance) would considerably reduce the speed of convergence. It might also be beneficial to work iteratively adding one curve at a time (with appropriate modifications to the loss to allow it to remain local). 	
\begin{figure}[!t]	
    \centering	
    	
    \begin{subfigure}[t]{.5\columnwidth}	
        \centering	
        \includegraphics[width=0.9\columnwidth]{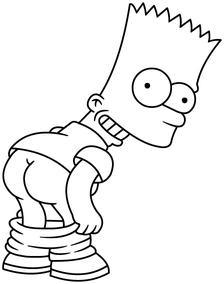}	
        \caption{Bart; target raster}	
        \label{fig:bart-raster}	
    \end{subfigure}\hfill%
    \begin{subfigure}[t]{.5\columnwidth}	
        \centering	
        \includegraphics[width=0.9\columnwidth]{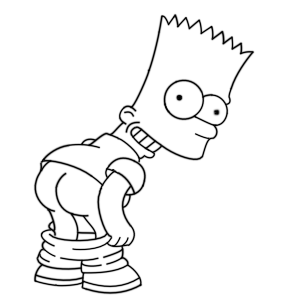}	
        \caption{Bart; generated vector}	
        \label{fig:bart-vector}	
    \end{subfigure}\hfill%
    \\ \vspace{0.5em}	
    \begin{subfigure}[t]{.5\columnwidth}	
        \centering	
        \includegraphics[width=0.8\columnwidth]{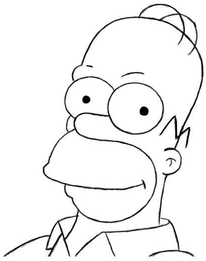}	
        \caption{Homer; target raster}	
        \label{fig:homer-raster}	
    \end{subfigure}\hfill%
    \begin{subfigure}[t]{.5\columnwidth}	
        \centering	
        \includegraphics[width=0.8\columnwidth]{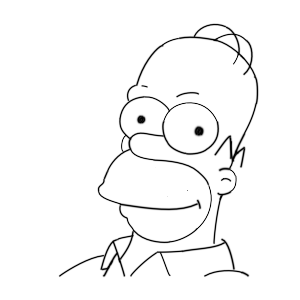}	
        \caption{Homer; generated vector}	
        \label{fig:homer-vector}	
    \end{subfigure}\hfill%
    \caption{Results of converting cartoon rasters to vectors using direct optimisation of Catmull-Rom Spline curve segments with randomly initialised curves and learned stroke width. 10000 iterations were performed, and for the first 5000 iterations any curve with near-zero stroke width was randomly reinitialised. Zoom-in to better view the differences and the details of the generated vector versions.}	
    \label{fig:simpsons}	
\end{figure}	
However, as demonstrated in \cref{fig:simpsons}, optimising randomly initialised curves (single segment Catmull-Rom splines) and their widths with the MSE loss works well for a cartoon raster to vector conversion task. The errors in these conversions are limited to missing small features --- for example in the ears. The only optimisation `tricks' required were to run for enough iterations (the \cref{fig:simpsons} images were allowed 10000 iterations, although they had converged before the end) and either using more lines than required, or using a smaller number and randomly re-initialising any that had widths learned to be zero (which become invisible) during the first half of the iteration limit. All of these details are only to overcome the fact that poorly initialised curves (ones far from any black pixel in the raster along their length) will naturally have strong gradients forcing them to be removed by reducing their stroke width to zero.	
\makeextravis{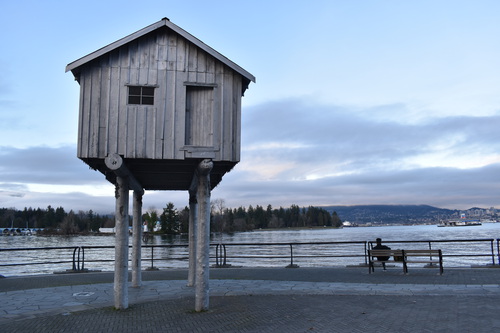}	
\makeextravis{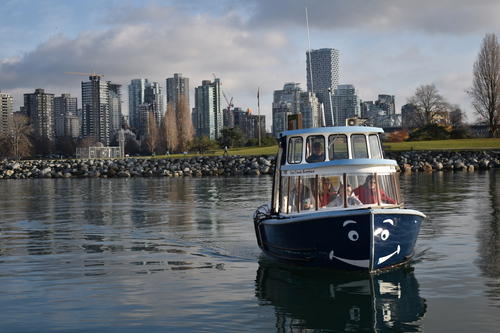}	
\makeextravis{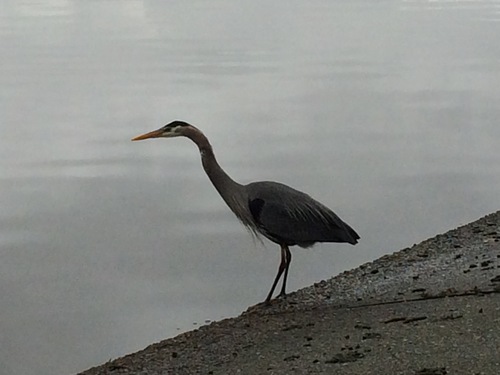}	
\makeextravis{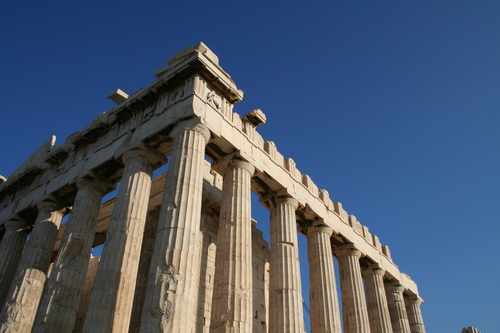}	
\makeextravis{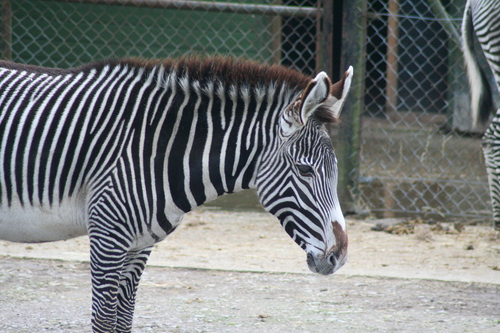}	
\makeextravis{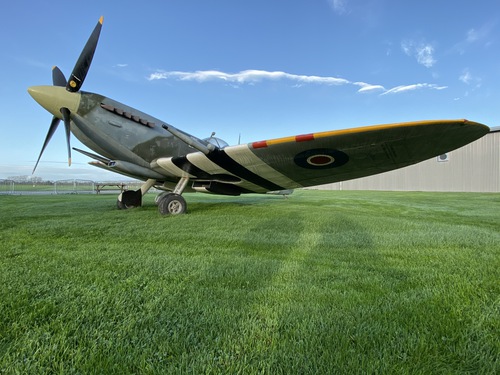}	
\makeextravis{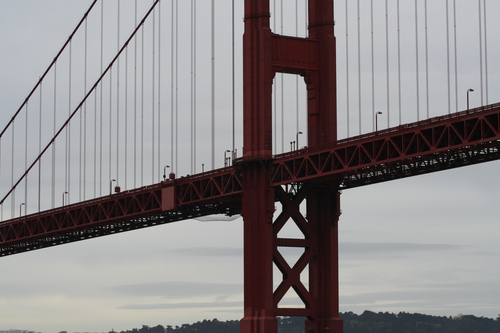}	
\makeextravis{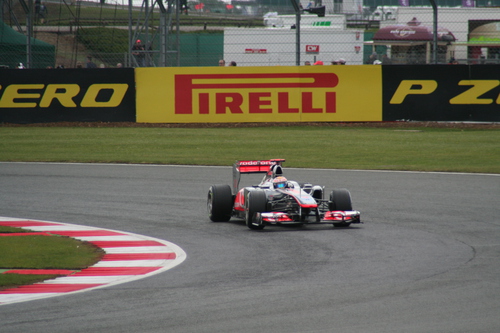}	
\makeextravis{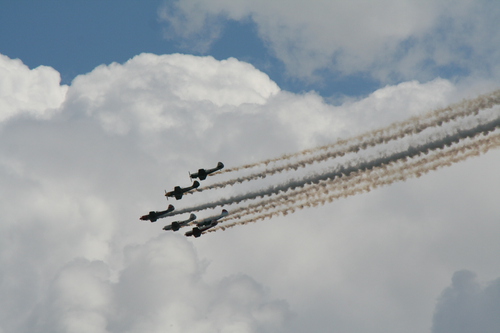}	
\makeextravis{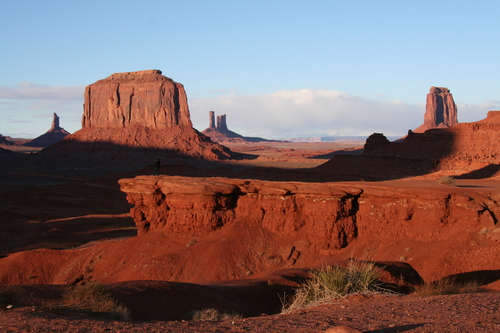}	
\makeextravis{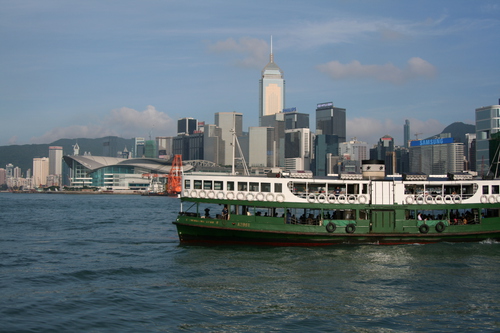}	
\makeextravis{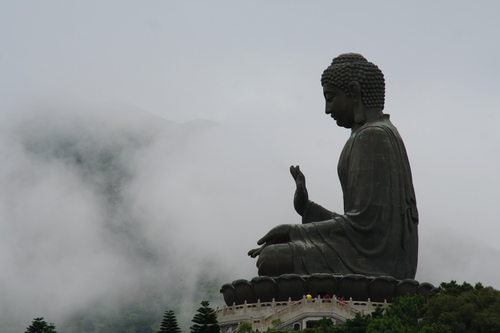}	
\makeextravis{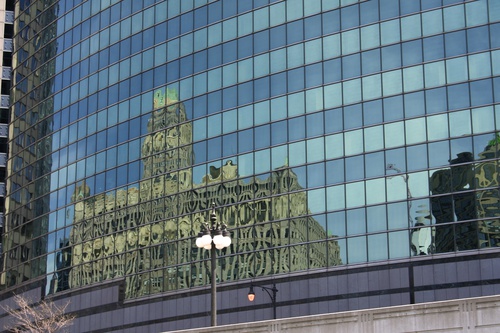}	
\makeextravis{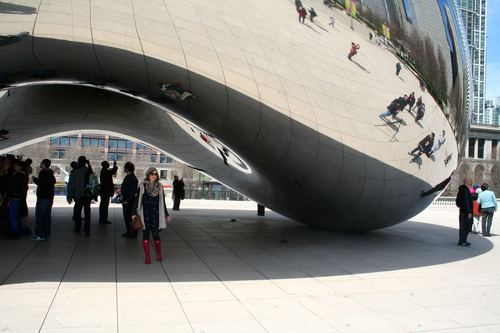}	

\section{Potential Challenges in Optimisation}\label{sup:optim-challenges}	
\begin{figure*}[!t]
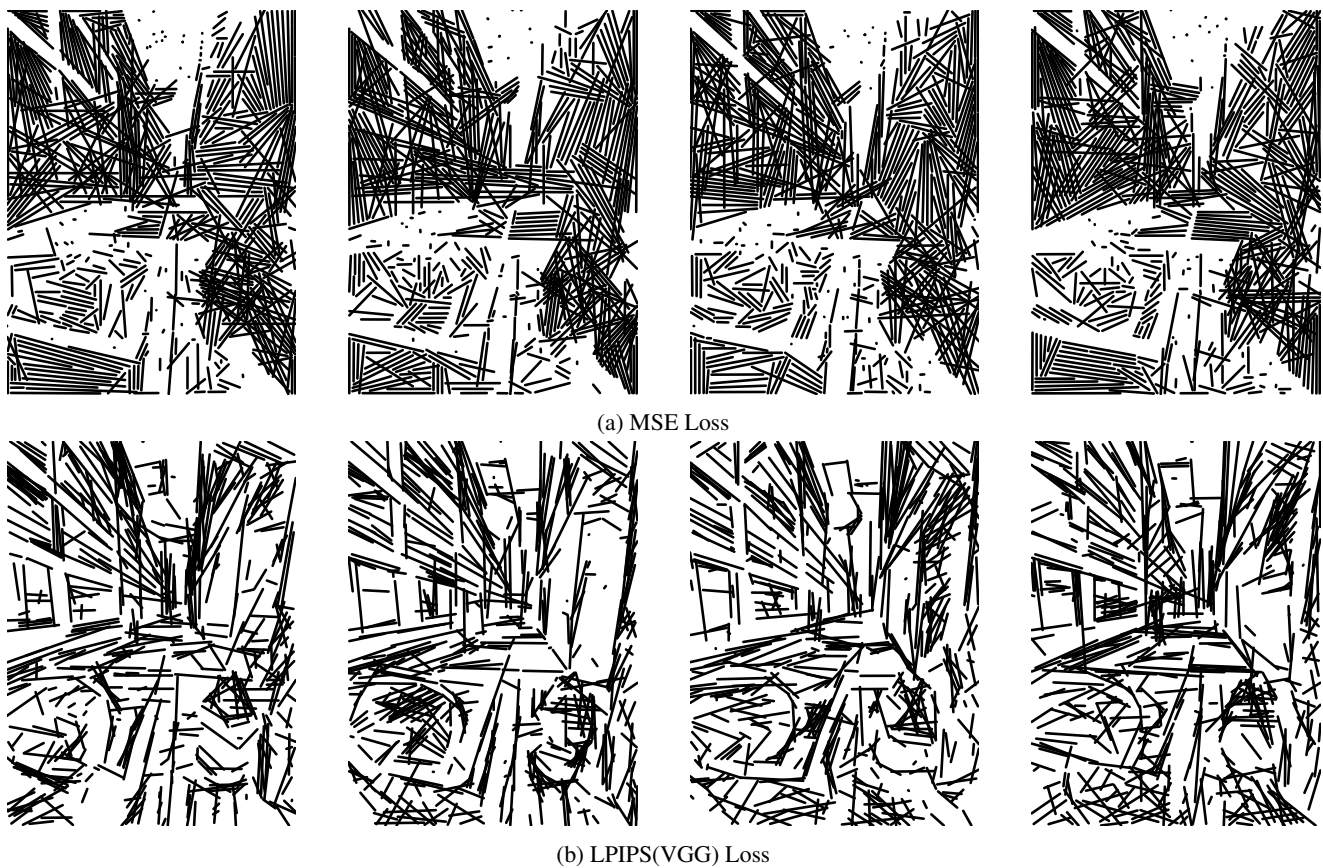
	
    \centering	
    \begin{subfigure}[t]{\textwidth}	
        \centering	
        \includegraphics[width=0.22\columnwidth]{images/randinits/imageopt-500lines-mse-seed0.pdf}\hfill	
        \includegraphics[width=0.22\columnwidth]{images/randinits/imageopt-500lines-mse-seed1.pdf}\hfill	
        \includegraphics[width=0.22\columnwidth]{images/randinits/imageopt-500lines-mse-seed2.pdf}\hfill	
        \includegraphics[width=0.22\columnwidth]{images/randinits/imageopt-500lines-mse-seed3.pdf}	
        \caption{MSE Loss}	
        \label{fig:mse-rand-init}	
    \end{subfigure}\hfill%
    \begin{subfigure}[t]{\textwidth}	
        \centering	
        \includegraphics[width=0.22\columnwidth]{images/randinits/imageopt-500lines-lpips-vgg-seed0.pdf}\hfill	
        \includegraphics[width=0.22\columnwidth]{images/randinits/imageopt-500lines-lpips-vgg-seed1.pdf}\hfill	
        \includegraphics[width=0.22\columnwidth]{images/randinits/imageopt-500lines-lpips-vgg-seed2.pdf}\hfill	
        \includegraphics[width=0.22\columnwidth]{images/randinits/imageopt-500lines-lpips-vgg-seed3.pdf}	
        \caption{LPIPS(VGG) Loss}	
        \label{fig:lpips-rand-init}	
    \end{subfigure}\hfill%
    \caption{Exploring the effect of different random initialisations. Four different random seeds were used for the initial lines, and the resultant images were created using MSE and LPIPS(VGG) loss. Images directly above/below each other correspond to the same random seed/initialisation.}	
    \label{fig:randinits}	
\end{figure*}

As mentioned in \cref{sec:optimisation} the loss landscape when using a differentiable rasterister is challenging because of factors such including considerable permutation symmetry from being able to draw strokes in either direction, as well as many local optima. In the case of direct optimisation, the resultant images can be very sensitive to initialisation; we discuss this further in the following subsection. For our autoencoder experiments we found no problems with training, however in the future we want to explore how different priors (e.g. preference for long strokes, preference to draw from left to right), could affect this.	
\paragraph{Effect of initialisation.}	
Simple losses like MSE are very sensitive to initialisation when used to optimise against a photo. As can be seen in \cref{fig:mse-rand-init}, initialisations can have a dramatic effect on the orientation of `shaded' sections of the resultant image. This is in itself not something to worry about as it just results in a \textit{artistically} different result, but the broad perceptual appearance of tone is still preserved. Different losses can overcome this sensitivity to an extent however. As can be seen in \cref{fig:lpips-rand-init}, using the LPIPS(VGG) loss for example always captures perceptually important directions in the resultant image, although the initialisation can still affect the rendition in more localised areas. 	

\section{Autotracing Model Architectures}
This section details the model architectures used for the autotracing autoencoders described in \cref{sec:ae} of the main paper.
\subsection{Encoders}
\label{sup:encoders}
A series of encoder networks were used for different datasets. In each encoder network architecture, one can control the dimensionality of the latent vector encoding (\verb|latent_size|).
\clearpage
\paragraph{MLP Encoder}
The encoder network used for MNIST experiments is a simple multi-layer perceptron with default $\texttt{hidden}=64$ and $\texttt{latent\_size}=64$:

\begin{verbatim}
Linear(28*28, hidden)
ReLU()
Linear(hidden, latent_size)
ReLU()
\end{verbatim}

\paragraph{CNN Encoder}
For resized Omniglot experiments ($28\times28$ pixel images), a convolutional encoder network with batch normalization was used:

\begin{verbatim}
Conv2d(in_channels=1, out_channels=64, 
    kernel_size=3, padding=1, stride=1)
BatchNorm2d(num_features=64)
ReLU()
Conv2d(in_channels=64, out_channels=64, 
    kernel_size=3, padding=1, stride=1)
BatchNorm2d(num_features=64)
ReLU()
Conv2d(in_channels=64, out_channels=64, 
    kernel_size=3, padding=1, stride=1)
BatchNorm2d(num_features=64)
ReLU()
Conv2d(in_channels=64, out_channels=64, 
    kernel_size=3, padding=1, stride=1)
BatchNorm2d(num_features=64)
ReLU()
AdaptiveAvgPool2d(output_size=8)
Flatten()
Linear(4096, latent_size)
\end{verbatim}

\paragraph{StrokeNet Agent Encoder} Finally, when comparing against StrokeNet \citep{zheng2018strokenet} on scaled MNIST ($256\times256$), we present results replicating \citeauthor{zheng2018strokenet}'s AgentCNN encoder\footnote{Code available at \url{https://github.com/vexilligera/strokenet}}.

\begin{verbatim}
Conv2d(in_channels=1, out_channels=16, 
    kernel_size=3, padding=1)
BatchNorm2d(num_features=16)
LeakyReLU(negative_slope=0.2)
Conv2d(in_channels=16, out_channels=16,
    kernel_size=3, padding=1)
BatchNorm2d(num_features=16)
LeakyReLU(negative_slope=0.2)
AvgPool2d(kernel_size=2)

Conv2d(in_channels=16, out_channels=32, 
    kernel_size=3, padding=1)
BatchNorm2d(num_features=32)
LeakyReLU(negative_slope=0.2)
Conv2d(in_channels=32, out_channels=32, 
    kernel_size=3, padding=1)
BatchNorm2d(num_features=32)
LeakyReLU(negative_slope=0.2)
AvgPool2d(kernel_size=2)

Conv2d(in_channels=32, out_channels=64, 
    kernel_size=3, padding=1)
BatchNorm2d(num_features=64)
LeakyReLU(negative_slope=0.2)
Conv2d(in_channels=64, out_channels=64, 
    kernel_size=3, padding=1)
BatchNorm2d(num_features=64)
LeakyReLU(negative_slope=0.2)
AvgPool2d(kernel_size=2)

Conv2d(in_channels=64, out_channels=128, 
    kernel_size=3, padding=1)
BatchNorm2d(num_features=128)
LeakyReLU(negative_slope=0.2)
Conv2d(in_channels=128, out_channels=128, 
    kernel_size=3, padding=1)
BatchNorm2d(num_features=128)
LeakyReLU(negative_slope=0.2)
AvgPool2d(kernel_size=2)

Conv2d(in_channels=128, out_channels=256, 
    kernel_size=3, padding=1)
BatchNorm2d(num_features=256)
LeakyReLU(negative_slope=0.2)
Conv2d(in_channels=256, out_channels=256, 
    kernel_size=3, padding=1)
BatchNorm2d(num_features=256)
LeakyReLU(negative_slope=0.2)
AvgPool2d(kernel_size=2)

Flatten()
\end{verbatim}

\subsection{Decoders}
\label{sup:decoders}
All our decoder networks, which provide different stroke parameterisations, have a common structure consisting of two linear layers followed by ReLU non-linear activation. For $28\times28$ pixel MNIST and Omniglot experiments, we used $\texttt{hidden1}=64$ and $\texttt{hidden2}=256$.
\begin{verbatim}
Linear(latent_size, hidden1)
ReLU()
Linear(hidden1, hidden2)
ReLU()
\end{verbatim}
This common structure is followed by a sub-network to produce stroke parameters; this is usually a single linear layer followed by a $\tanh$ function. Specific details for the chosen type of primitive parameterisation is as follows:

\paragraph{Line.}
Line decoder outputs the start and end coordinates of \texttt{nlines} segments. The default number of lines used for MNIST is 5.
\begin{verbatim}
Linear(hidden2, nlines * 4)
Tanh()
\end{verbatim}

\paragraph{PolyLine.} 
This allows us to decode the stroke data to a sequence of consecutive points (each defined as $\bm{p}=[p_x,p_y]$). Default value of \texttt{npoints} is 16, but it can be varied.
\begin{verbatim}
Linear(hidden2, npoints * 2)
Tanh()
\end{verbatim}

\paragraph{PolyConnect.}
Similar to \textit{PolyLine}, but instead of decoding to a sequence of consecutive points, it outputs a set of points joined together by a learned connection matrix. The network computing \texttt{npoints} 2d coordinates is the same as in \textit{PolyLine} and the sub-network computing the upper triangular part of the connection matrix is:
\begin{verbatim}
Linear(hidden2, nlines)
Sigmoid()
\end{verbatim}
\noindent where \texttt{nlines} is computed as
\begin{verbatim}
int((npoints ** 2 + npoints) / 2)
\end{verbatim}
if we allow single points to be drawn (\ie compute the diagonal of the connection matrix), and as follows, otherwise:
\begin{verbatim}
int(npoints * (npoints - 1) / 2)
\end{verbatim}
All possible combinations of lines formed between the set of \texttt{npoints} are rasterised and shaded by the appropriate connection weight before composition.

\paragraph{CRS.} Decoder that parametrises stroke data as Catmull-Rom splines with default of $\texttt{nlines}=1$ and $\texttt{npoints}=16$ control points:
\begin{verbatim}
Linear(hidden2, nlines * npoints * 2),
Tanh()
\end{verbatim}

\paragraph{B\'ezier.} We can also choose to parametrise strokes as B\'ezier curved lines (default $\texttt{nlines}=5$) and can specify the number of \texttt{segments} (default is 1).
\begin{verbatim}
Linear(hidden2, 2 * npoints * nlines)
Tanh()   
\end{verbatim}
where the number of control points is computed based on the number of segments:
\begin{verbatim}
npoints = (4 + (segments - 1) * 3)
\end{verbatim}

Note that this allows for a connected path which isn't necessarily smooth as each segment has independent control points. It is possible to formulate a version which is smooth and has fewer ($4 + (\texttt{segments} - 1) \times 2$) control points.

\paragraph{B\'ezierConnect.} Following the same pattern as \textit{PolyConnect}, this decodes stroke data to a set of control points for B\'ezier curves. The connection matrix is learned using a network as described in the paragraph \textit{PolyConnect} and each point corresponds to both a curve's end point and corresponding control point to allow for smooth curve segments.

\subsection{Recurrent Decoders}
We implemented \citet{zheng2018strokenet}'s recurrent model that at each time step uses two separate CNN networks, one to encode the target image and one for the previous frame. The vector encodings are concatenated, decoded to `stroke data' and a new stroke is rendered. The new stroke is then overlaid on the previous frame. However, this approach proved is computationally expensive and difficult to train well.

Instead, we used a GRU-based RNN which initially starts with a projection of a zeroed input and the target image's vector encoding as its initial hidden state. The RNN decoder architecture is as follows:

\begin{verbatim}
Linear(output_size, latent_size)
ReLU()
GRU(latent_size, latent_size)
Linear(latent_size, output_size)
\end{verbatim}
where \texttt{output\_size} can be modified depending on the chosen type of primitive parametrisation (\eg a B\'ezier curve has 4 control points, hence $\texttt{output\_size}=8$). The GRU model is both trained and evaluated for a predefined number of time steps (3 in all our experiments), corresponding to the number of independent strokes produced.

\section{MNIST Reconstructions}
\label{sup:MNISTrecon}

\begin{figure*}[t]
    \centering
    \begin{subfigure}[t]{.25\textwidth}
        \centering
        \includegraphics[trim=0 28 0 0,clip,width=0.9\columnwidth]{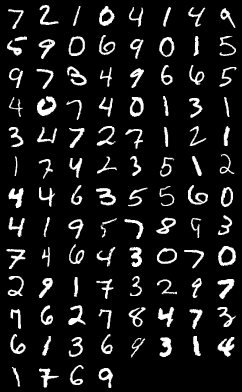}
        \caption{Test Samples}
        \label{fig:mnistsamples}
    \end{subfigure}\hfill%
    \begin{subfigure}[t]{.25\textwidth}
        \centering
        \includegraphics[trim=0 28 0 0,clip,width=0.9\columnwidth]{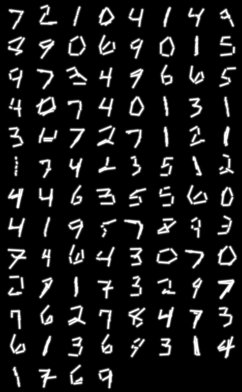}
        \caption{Lines(L=5)}
        \label{fig:linerec}
    \end{subfigure}\hfill%
    \begin{subfigure}[t]{.25\textwidth}
        \centering
        \includegraphics[trim=0 28 0 0,clip,width=0.9\columnwidth]{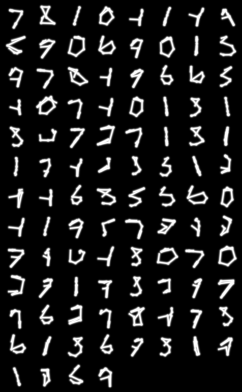}
        \caption{PolyLine(P=8)}
        \label{fig:polyline8pts}
    \end{subfigure}\hfill%
    \begin{subfigure}[t]{.25\textwidth}
        \centering
        \includegraphics[trim=0 28 0 0,clip,width=0.9\columnwidth]{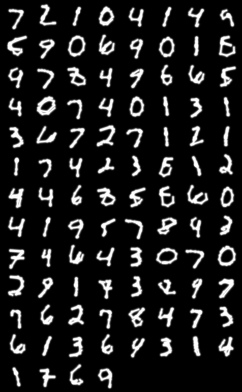}
        \caption{PolyLine(P=16)}
        \label{fig:polyline}
    \end{subfigure}\hfill%
    
    \vspace{0.5em}
    
    \centering
    \begin{subfigure}[t]{.25\textwidth}
        \centering
        \includegraphics[trim=0 28 0 0,clip,width=0.9\columnwidth]{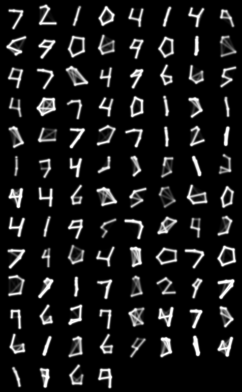}
        \caption{PolyConnect(P=5)}
        \label{fig:polyconn5pts}
    \end{subfigure}\hfill%
    \begin{subfigure}[t]{.25\textwidth}
        \centering
        \includegraphics[trim=0 28 0 0,clip,width=0.9\columnwidth]{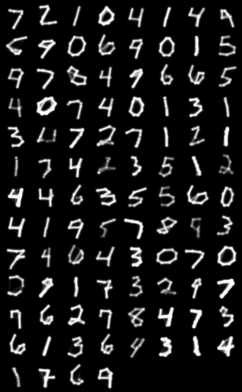}
        \caption{PolyConnect(P=8)}
        \label{fig:polyconn8pts}
    \end{subfigure}\hfill%
    \begin{subfigure}[t]{.25\textwidth}
        \centering
        \includegraphics[trim=0 28 0 0,clip,width=0.9\columnwidth]{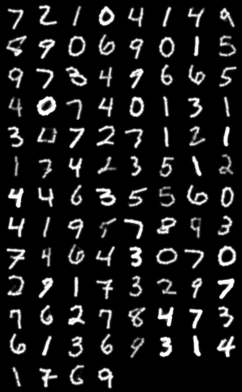}
        \caption{PolyConnect(P=16)}
        \label{fig:polyconn16pts}
    \end{subfigure}\hfill%
    \begin{subfigure}[t]{.25\textwidth}
        \centering
        \includegraphics[trim=0 28 0 0,clip,width=0.9\columnwidth]{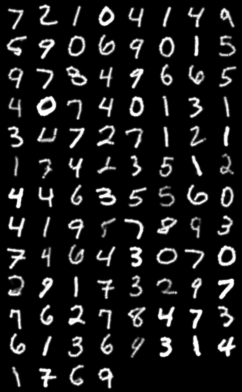}
        \caption{PolyConnect(P=32)}
        \label{fig:polyconn32pts}
    \end{subfigure}\hfill%

  \caption{MNIST test set samples and reconstructions using different parameterisations of `stroke data': Lines, PolyLine (\ie a series of consecutive (P)oints) and PolyConnect (a set of 2d (P)oints joined by a learned connection matrix). }
  \label{fig:mnistlinepointsrec}
\end{figure*}

\begin{figure*}[t]
    \centering
    \begin{subfigure}[t]{.25\textwidth}
        \centering
        \includegraphics[trim=0 28 0 0,clip,width=0.9\columnwidth]{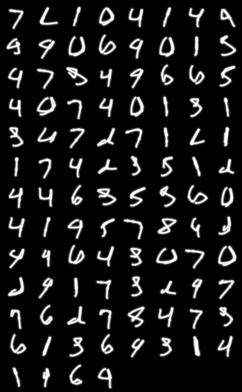}
        \caption{CRS(L=1, P=8)}
        \label{fig:crs1l8pts}
    \end{subfigure}\hfill%
    \begin{subfigure}[t]{.25\textwidth}
        \centering
        \includegraphics[trim=0 28 0 0,clip,width=0.9\columnwidth]{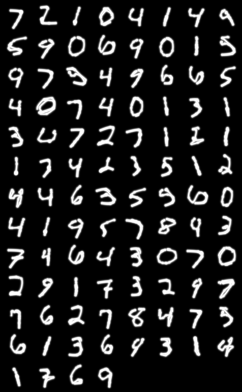}
        \caption{CRS(L=1, P=16)}
        \label{fig:crsrecon}
    \end{subfigure}\hfill%
    \begin{subfigure}[t]{.25\textwidth}
        \centering
        \includegraphics[trim=0 28 0 0,clip,width=0.9\columnwidth]{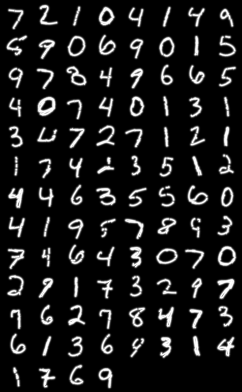}
        \caption{B\'ezier(L=5, S=1)}
        \label{fig:bezier}
    \end{subfigure}\hfill%
    \begin{subfigure}[t]{.25\textwidth}
        \centering
        \includegraphics[trim=0 28 0 0,clip,width=0.9\columnwidth]{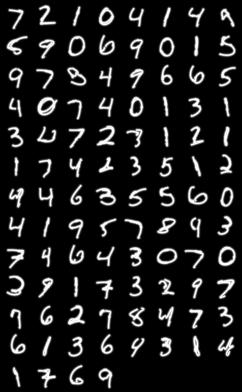}
        \caption{B\'ezier(L=2, S=2)}
        \label{fig:bez2l2s}
    \end{subfigure}\hfill%
    
    \vspace{0.5em}
    
    \centering
    \begin{subfigure}[t]{.25\textwidth}
        \centering
        \includegraphics[trim=0 28 0 0,clip,width=0.9\columnwidth]{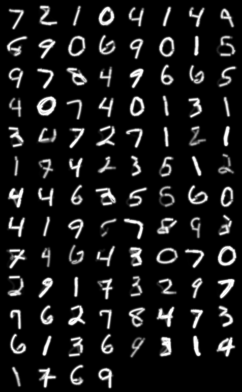}
        \caption{B\'ezierConnect(P=5)}
        \label{fig:bezconn5pts}
    \end{subfigure}\hfill%
    \begin{subfigure}[t]{.25\textwidth}
        \centering
        \includegraphics[trim=0 28 0 0,clip,width=0.9\columnwidth]{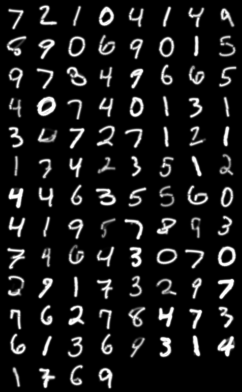}
        \caption{B\'ezierConnect(P=8)}
        \label{fig:bezconn8pts}
    \end{subfigure}\hfill%
    \begin{subfigure}[t]{.25\textwidth}
        \centering
        \includegraphics[trim=0 28 0 0,clip,width=0.9\columnwidth]{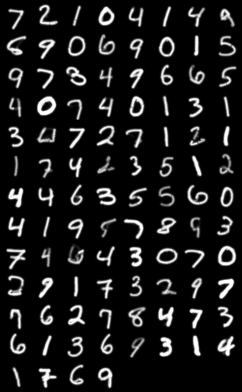}
        \caption{B\'ezierConnect(P=16)}
        \label{fig:bezconn16pts}
    \end{subfigure}\hfill%
    \begin{subfigure}[t]{.25\textwidth}
        \centering
        \includegraphics[trim=0 28 0 0,clip,width=0.9\columnwidth]{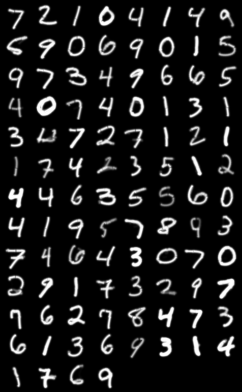}
        \caption{B\'ezierConnect(P=32)}
        \label{fig:bezconn32pts}
    \end{subfigure}\hfill%
    
  \caption{MNIST test set reconstructions (of samples in \cref{fig:mnistsamples}) with curves parametrised as Catmull-Rom splines (CRS) and B\'ezier curves (B\'ezier and B\'ezierConnect). In both CRS and B\'ezier Decoders, we can vary the number of (L)ines, (P)oints and, respectively, (S)egments. B\'ezierConnect allows control over the (P)oints joined by the learned connection matrix.}
  \label{fig:mnistcurverecons}
\end{figure*}

\Cref{fig:mnistlinepointsrec,fig:mnistcurverecons} illustrate the effect of different stroke parameterisations of the MNIST dataset. Varying the number of (L)ines, (S)egments, and (P)oints and introducing a learned connection matrix between them leads to distinct approaches to drawing. As depicted in \cref{fig:polyconn16pts,fig:bezconn16pts}, \textit{Connect} models produce the closest reconstructions. Likewise, parameterisation using simple B\'ezier curves (\cref{fig:bezier}) leads to convincing results.  

\section{Omniglot ($28\times28$ pixels) Comparison} 
\label{sup:omniglotresults}
\begin{figure*}[t]	
    \centering	
    \begin{subfigure}[t]{.16\textwidth}	
        \centering	
        \includegraphics[width=0.9\columnwidth]{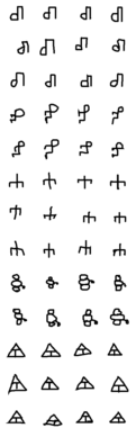}	
        \caption{Test Samples}	
    \end{subfigure}\hfill%
    \begin{subfigure}[t]{.16\textwidth}	
        \centering	
        \includegraphics[width=0.885\columnwidth]{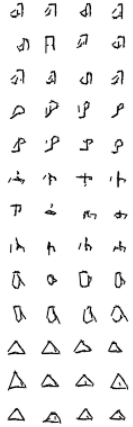}	
        \caption{Lines(L=10)}	
    \end{subfigure}\hfill%
    \begin{subfigure}[t]{.16\textwidth}	
        \centering	
        \includegraphics[width=0.915\columnwidth]{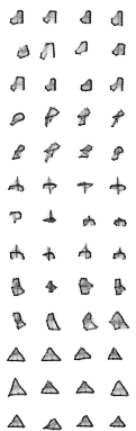}	
        \caption{PolyConn(P=16)}	
    \end{subfigure}\hfill%
    \begin{subfigure}[t]{.16\textwidth}	
        \centering	
        \includegraphics[width=0.88\columnwidth]{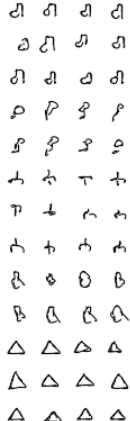}	
        \caption{B\'ezier(L=5,S=1)}	
    \end{subfigure}\hfill%
    \begin{subfigure}[t]{.16\textwidth}	
        \centering	
        \includegraphics[width=0.9\columnwidth]{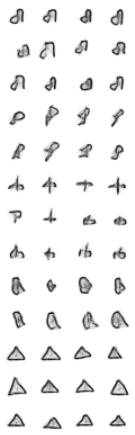}	
        \caption{B\'ezierConn(P=16)}	
    \end{subfigure}\hfill%
    \begin{subfigure}[t]{.16\textwidth}	
        \centering	
        \includegraphics[width=0.9\columnwidth]{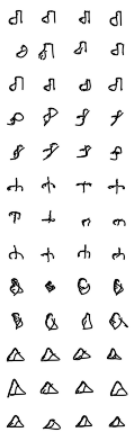}	
        \caption{RNNB\'ezier(St=10)}	
    \end{subfigure}\hfill%
  \caption{Omniglot test set samples and reconstructions using different parameterisations of ‘stroke data'.}	
  \label{fig:omniglot-supplementary-images}	
\end{figure*}

\Cref{Table:OmniglotComparison} shows the effect of different parameterisations on Omniglot dataset. All the models demonstrate reasonable generalisation to the test dataset (as measured by MSE) even though the test alphabets are completely disjoint from the training/validation ones.  Reconstructions of models with different parametrisations are shown in \cref{fig:omniglot-supplementary-images}. B\'ezier curves work particularly well, although we note that they do appear to struggle with forming dots (for example in the Braille alphabet which can found in the training/validation sets).
\begin{table}[H]
\begin{center}
\begin{tabular}{lcccccc}	
\toprule
Decoder & St & \#P & \#S &\#L & Val & Test\\
\midrule
Line & 1 & 20 & 1& 10 & 0.0189 & 0.0223 \\
PolyConnect & 1 & 16 &  - & - &0.0127 & 0.0151\\
B\'ezier & 1 & 20 & 1 & 5 & 0.0158 & 0.0194 \\
B\'ezierConnect & 1 & 16 & - & - & 0.0117 & 0.0144\\
RNNB\'ezier & 10 & 16 & 1 & 1 & 0.0152 & 0.0181\\
B\'ezier* & 1 & 50 & 3 & 5 & 0.0091 & 0.0118\\

\bottomrule
\end{tabular}
\end{center}
\caption{Omniglot validation and test MSE for models constructed with different parameterisations and architecture (\ie recurrent vs single-(St)ep). B\'ezier* corresponds to the model whose reconstructions were shown in ~\cref{fig:omniglotsketches} and has $\texttt{hidden1}=512$ and $\texttt{hidden2}=1024$.}
\label{Table:OmniglotComparison}
\end{table}

\section{StrokeNet ScaledMNIST Comparison}
\label{sup:strokenet}
The StrokeNet paper~\citep{zheng2018strokenet} describes an evaluation of the model on scaled-up MNIST characters by comparing performance against a CNN-based classifier trained on the scaled images, and then evaluated on the reconstructions. The paper implies that the MNIST characters were just re-sampled to 256x256, however from analysis of the source code it can be determined that the scaling procedure was to: resize the 28x28 characters to 120x120 using bilinear interpolation, pad the 120x120 images to 256x256, and change the contrast by multiplying pixels by 0.6. Although the original rationale for these choices is unclear, we follow exactly the same procedure for our experiments.

The structure of the classifier model in the paper is not described beyond it being convolutional with 5-layers, and no code for this aspect of the experiments was provided. We thus chose to implement our own classifier as follows:

\begin{verbatim}
Conv2d(in_channels=1, out_channels=30, 
    kernel_size=5, padding=0, stride=1)
ReLU()
Conv2d(in_channels=30, out_channels=15, 
    kernel_size=5, padding=0, stride=1)
ReLU()
Linear(6000, 128)
ReLU()
Linear(128, 50)
ReLU()
Linear(50, 10)
\end{verbatim}

\noindent   We did not use any form of regularisation or dropout during training. The network was trained for 10 epochs using the Adam optimiser with a learning rate of 0.001 and PyTorch's \verb|CrossEntropyLoss| which incorporates the Softmax activation. This network performs considerably better than the results presented in the original paper on the raw scaled MNIST test dataset (originally reported accuracy is 90.82\%, whereas the above network achieves 98.58\%). To compute the performance of the StrokeNet paper with our classification network we take the pretrained model weights provided by the StrokeNet authors and use them to generate reconstructions of the scaled MNIST test set, which are then fed to the classifier network to make predictions from. Again we found considerably higher performance than was originally reported, as detailed in the main paper.

\section{Additional Autoencoders Results}\label{sup:additional-AE-results}	
We provide results of autotracing autoencoders tested on additional datasets: Japanese handwritten characters \citep{clanuwat2018deep} and human quick drawings.	
\subsection{KMNIST ($28\times28$ pixels)} 	
\label{sup:KMNISTresults}	
\begin{figure*}[t]	
    \centering	
    \begin{subfigure}[t]{.25\textwidth}	
        \centering	
        \includegraphics[trim=0 32 0 0,clip,width=0.9\columnwidth]{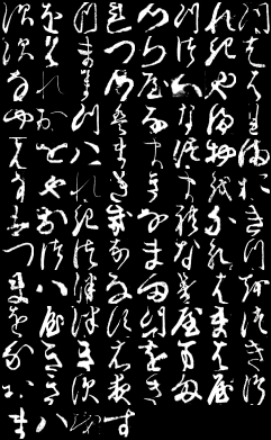}	
        \caption{Test Samples}	
    \end{subfigure}\hfill%
    \begin{subfigure}[t]{.25\textwidth}	
        \centering	
        \includegraphics[trim=0 32 0 0,clip,width=0.9\columnwidth]{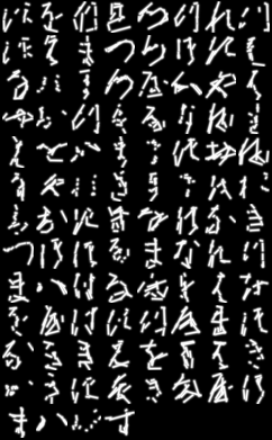}	
        \caption{Lines(L=10)}	
    \end{subfigure}\hfill%
    \begin{subfigure}[t]{.25\textwidth}	
        \centering	
        \includegraphics[trim=0 32 0 0,clip,width=0.9\columnwidth]{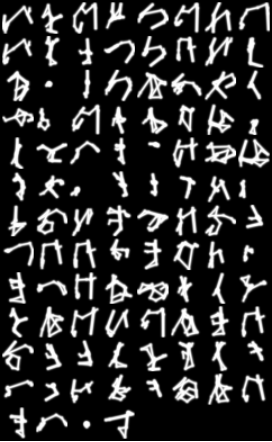}	
        \caption{PolyLine(P=16)}	
    \end{subfigure}\hfill%
    \begin{subfigure}[t]{.25\textwidth}	
        \centering	
        \includegraphics[trim=0 32 0 0,clip,width=0.9\columnwidth]{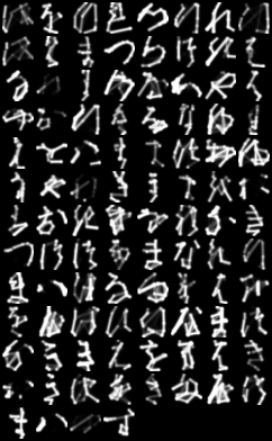}	
        \caption{PolyConnect(P=16)}	
    \end{subfigure}\hfill%
    	
    \vspace{0.5em}	
    	
    \centering	
    \begin{subfigure}[t]{.25\textwidth}	
        \centering	
        \includegraphics[trim=0 32 0 0,clip,width=0.9\columnwidth]{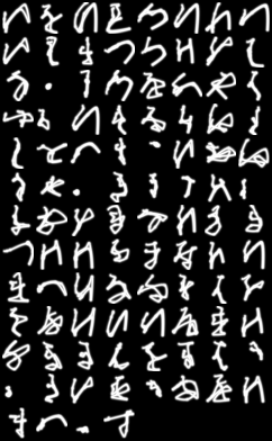}	
        \caption{CRS(L=1, P=16)}	
    \end{subfigure}\hfill%
    \begin{subfigure}[t]{.25\textwidth}	
        \centering	
        \includegraphics[trim=0 32 0 0,clip,width=0.9\columnwidth]{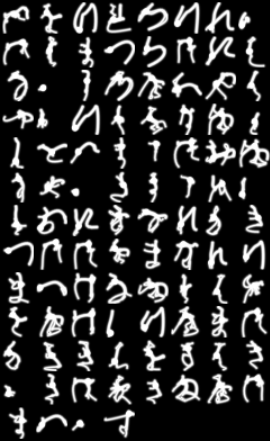}	
        \caption{B\'ezier(L=7, S=10)}	
    \end{subfigure}\hfill%
    \begin{subfigure}[t]{.25\textwidth}	
        \centering	
        \includegraphics[trim=0 32 0 0,clip,width=0.9\columnwidth]{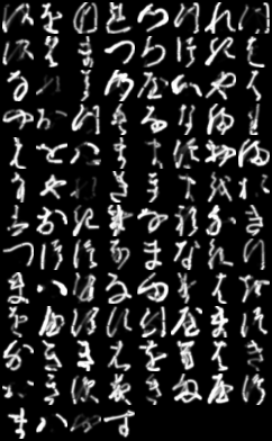}	
        \caption{B\'ezierConnect(P=16)}	
        \label{fig:kmnist-bezierconnect}	
    \end{subfigure}\hfill%
    \begin{subfigure}[t]{.25\textwidth}	
        \centering	
        \includegraphics[trim=0 32 0 0,clip,width=0.9\columnwidth]{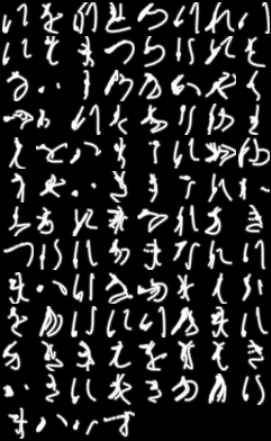}	
        \caption{RNNB\'ezier(St=10)}	
    \end{subfigure}\hfill%
  \caption{KMNIST test set samples and reconstructions using different parameterisations of `stroke data'.}	
  \label{fig:kmnist_sketches}	
\end{figure*}

\vspace{-0.5em}	
\begin{table}[H]	
\begin{center}	
\begin{tabular}{lcccccc}	
\toprule	
Decoder & St & \#P & \#S &\#L & Test & Acc. \%\\	
\midrule	
Line & 1 & 20 & 1 & 10 & 0.0431 & 87.2 \\	
PolyLine & 1 & 16 & 15 & 1 & 0.0654 & 75.06\\	
PolyConnect & 1 & 16 &  - & - & 0.0282 & 89.09\\	
CRS & 1 & 16 & 14 & 1 & 0.0635 & 76.2\\	
B\'ezier & 1 & 217 & 10 & 7 & 0.061 & 82.07 \\	
B\'ezierConnect & 1 & 16 & - & - & 0.0249 & 90.15\\	
RNNB\'ezier & 10 & 16 & 1 & 1 & 0.0496 & 80.19\\	
\bottomrule	
\end{tabular}	
\end{center}	
\vspace{-1em}	
\caption{KMNIST test MSE and classification accuracy (with a classifier trained on the un-encoded training set) for models constructed with different parameterisations.}	
\label{Table:KMNISTComparison}	
\end{table}	
\Cref{Table:KMNISTComparison} shows a comparison between different parametrisations performed on KMNIST \citep{clanuwat2018deep}, the Japanese Hiragana dataset. We provide test MSE and the classification accuracy of the drawn sketches. Samples of test reconstructions using different decoders are shown in \cref{fig:kmnist_sketches}. The \textit{B\'ezierConnect} model reaches the highest accuracy and creates the closest reconstructions as shown in \cref{fig:kmnist-bezierconnect}.	
\subsection{QuickDraw ($128\times128$ pixels)} 	
\label{sup:QuickDrawresults}	
\begin{figure*}[t]	
    \centering	
    \begin{subfigure}[t]{0.25\textwidth}	
        \centering	
        \includegraphics[width=0.9\columnwidth]{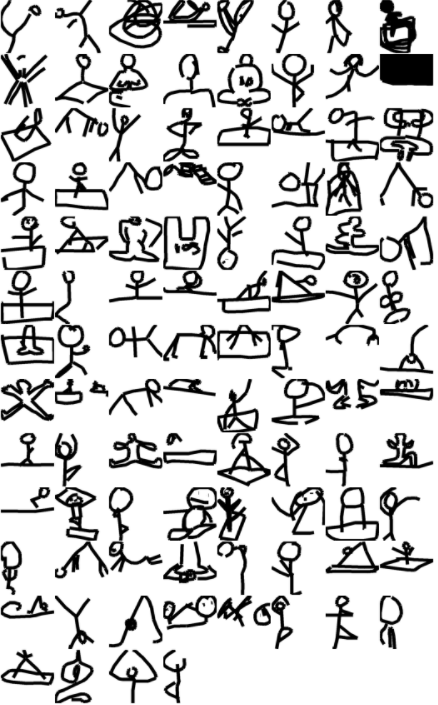}	
        \caption{Test Samples}	
    \end{subfigure}\hfill%
    \begin{subfigure}[t]{0.25\textwidth}	
        \centering	
        \includegraphics[width=0.9\columnwidth]{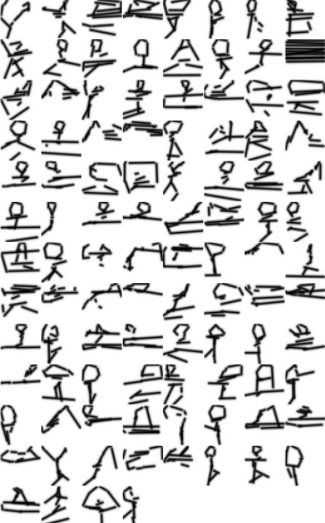}	
        \caption{Line(L=10)}	
    \end{subfigure}\hfill%
    \begin{subfigure}[t]{0.25\textwidth}	
        \centering	
        \includegraphics[width=0.9\columnwidth]{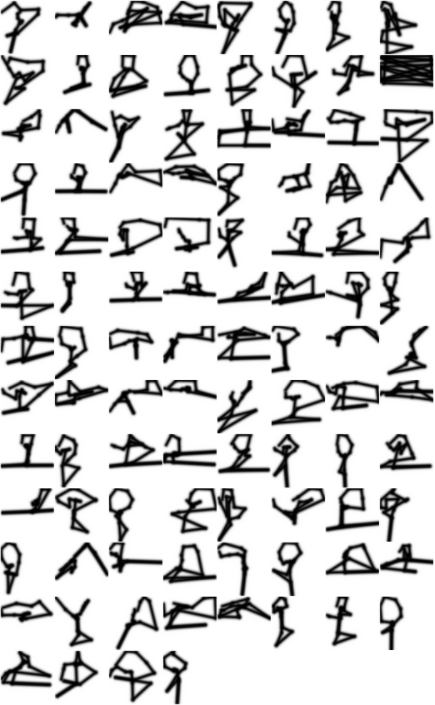}	
        \caption{PolyLine(P=16)}	
    \end{subfigure}\hfill%
    \begin{subfigure}[t]{0.25\textwidth}	
        \centering	
        \includegraphics[width=0.9\columnwidth]{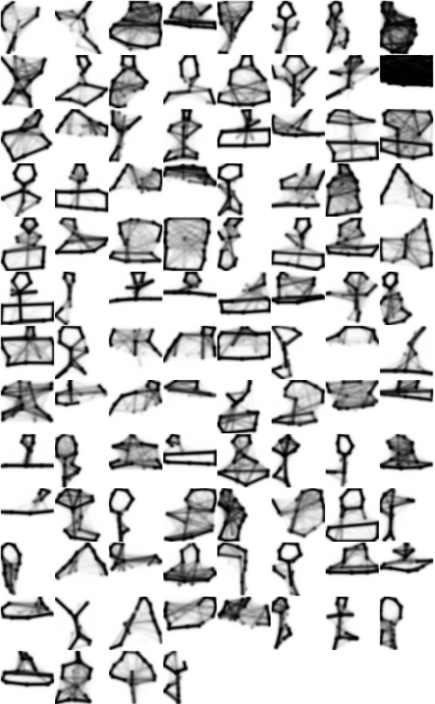}	
        \caption{PolyConnect(P=16)}	
        \label{fig:qd-polyconnect}	
    \end{subfigure}\hfill%
    	
    \vspace{0.5em}	
    	
    \centering	
    \begin{subfigure}[t]{0.25\textwidth}	
        \centering	
        \includegraphics[width=0.9\columnwidth]{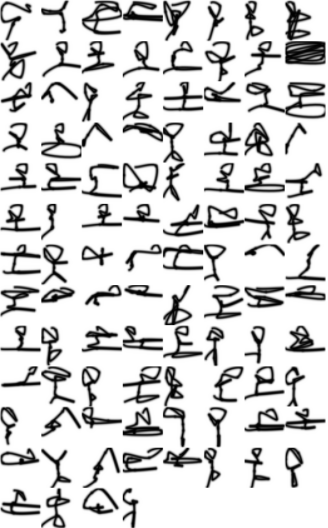}	
        \caption{CRS(L=1, P=16)}	
    \end{subfigure}\hfill%
    \begin{subfigure}[t]{0.25\textwidth}	
        \centering	
        \includegraphics[width=0.9\columnwidth]{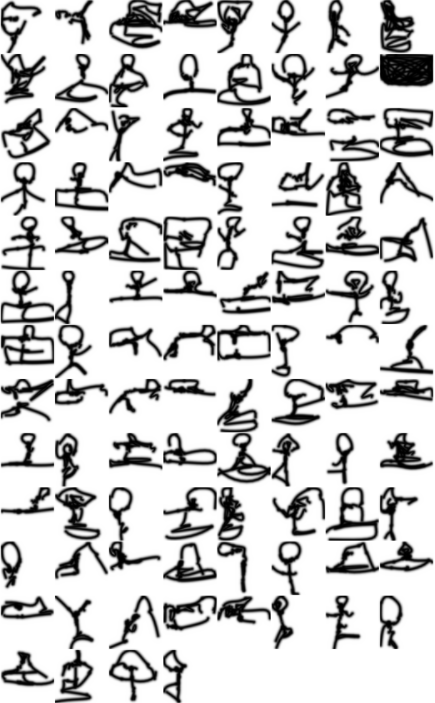}	
        \caption{B\'ezier(L=5, S=3)}	
    \end{subfigure}\hfill%
    \begin{subfigure}[t]{0.25\textwidth}	
        \centering	
        \includegraphics[width=0.9\columnwidth]{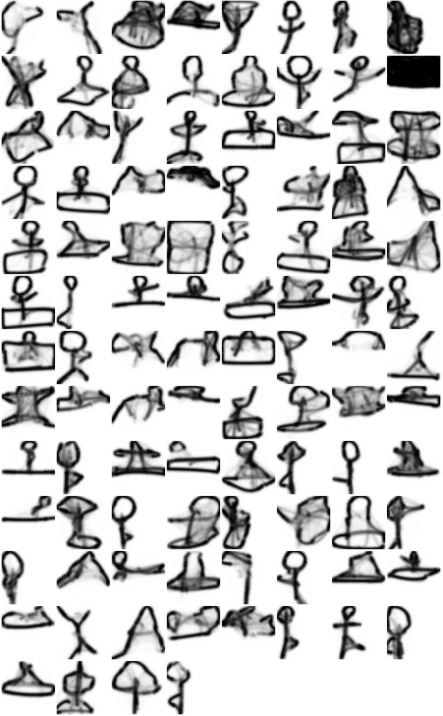}	
        \caption{B\'ezierConnect(P=16)}	
        \label{fig:qd-bezierconnect}	
    \end{subfigure}\hfill%
    \begin{subfigure}[t]{0.25\textwidth}	
        \centering	
        \includegraphics[width=0.9\columnwidth]{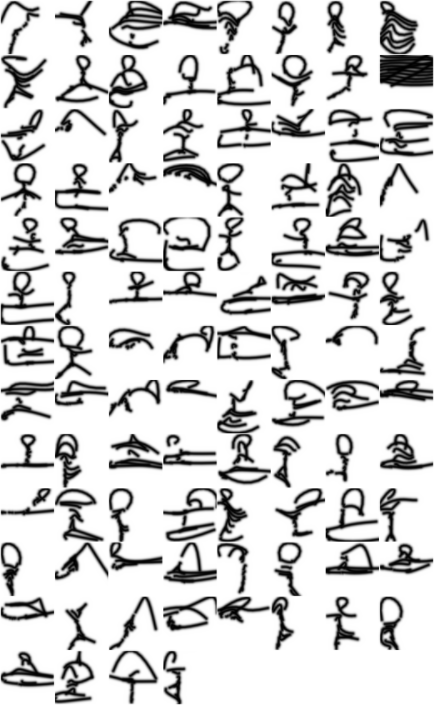}	
        \caption{RNNB\'ezier(St=10)}	
    \end{subfigure}\hfill%
  \caption{QuickDraw test set samples and reconstructions using different parameterisations of `stroke data'. Learning the connections between points leads to the most similar reconstructions (\cref{fig:qd-polyconnect,fig:qd-bezierconnect}).}	
  \label{fig:quickdraw}	
\end{figure*}

Next, we presents results of the autotracing experiment run on the Yoga class of QuickDraw\footnote{\url{https://github.com/googlecreativelab/quickdraw-dataset}}, a 50 million human drawing dataset across 345 image categories. We used 70000 doodles of yoga poses and split them so that the test, validation and train subsets were disjoint. \Cref{Table:QDComparison} shows validation and test MSE for different parametrisations. \Cref{fig:quickdraw} illustrates reconstructions of test samples for the different models. As seen before, learning the connections between points leads to the best results and produces the most similar reconstructions (\cref{fig:qd-bezierconnect,fig:qd-polyconnect}).	
\begin{table}[H]	
\begin{center}	
\begin{tabular}{lcccccc}	
\toprule	
Decoder & St & \#P & \#S &\#L & Val & Test\\	
\midrule	
Line & 1 & 20 & 1 & 10 & 0.086 & 0.080 \\	
PolyLine & 1 & 16 & 15 & 1 & 0.101 & 0.092\\	
PolyConnect & 1 & 16 &  - & - & 0.063 & 0.062\\	
CRS & 1 & 16 & 14 & 1 & 0.100 & 0.091\\	
B\'ezier & 1 & 50 & 3 & 5 & 0.0766 & 0.070 \\	
B\'ezierConnect & 1 & 16 & - & - & 0.049 & 0.048\\	
RNNB\'ezier & 10 & 16 & 1 & 1 & 0.0844 & 0.076\\	
\bottomrule	
\end{tabular}	
\end{center}	
\vspace{-1em}	
\caption{QuickDraw validation and test MSE for models constructed with different parameterisations.}	
\label{Table:QDComparison}	
\end{table}	

\end{document}